\documentclass[conference]{IEEEtran}
\IEEEoverridecommandlockouts

\usepackage{makecell}
\usepackage{subcaption}
\usepackage{pgf}
\usepackage{url}
\usepackage{hyperref}

\usepackage{comment}
\usepackage{cite}
\usepackage{amsmath,amssymb,amsfonts}
\usepackage{algorithmic}
\usepackage{graphicx}
\usepackage{textcomp}
\usepackage{xcolor}
\usepackage[font={footnotesize}]{caption}

\def\BibTeX{{\rm B\kern-.05em{\sc i\kern-.025em b}\kern-.08em
    T\kern-.1667em\lower.7ex\hbox{E}\kern-.125emX}}
\begin{document}

\title{\LARGE \bf \vspace{5mm}
Opti-Acoustic Scene Reconstruction in Highly Turbid \\Underwater Environments\\

\vspace{-4mm}

\thanks{{\footnotesize{$^{1}$I. Collado-Gonzalez, P. Szenher, and B. Englot are with Stevens Inst. of Technology, Hoboken, NJ, USA, \{\texttt{icollado}, \texttt{pszenher}, \texttt{benglot}\}\texttt{@stevens.edu}. $^{2}$J. McConnell is with the U.S. Naval Academy, Annapolis, MD, USA, \texttt{jmcconne@usna.edu}. \textbf{This research was supported in part by NSF grant IIS-1652064, USDA-NIFA grant 2021-67022-35977, and ONR grant N00014-24-1-2522.}}}}
}

\author{\IEEEauthorblockN{Ivana Collado-Gonzalez$^{1}$, John McConnell$^{2}$, Paul Szenher$^{1}$, and Brendan Englot$^{1}$}
}

\maketitle

\begin{abstract}
Scene reconstruction is an essential capability for underwater robots navigating in close proximity to structures. Monocular vision-based reconstruction methods 
are unreliable in turbid waters and lack depth scale information. Sonars are robust to turbid water and non-uniform lighting conditions, however, they have low resolution and elevation ambiguity. This work proposes a real-time opti-acoustic scene reconstruction method that is specially optimized to work in turbid water. Our strategy avoids having to identify point features in visual data and instead identifies regions of interest in the data. We then match relevant regions in the image to corresponding sonar data. A reconstruction is obtained by leveraging range data from the sonar and elevation data from the camera image. Experimental comparisons against other vision-based and sonar-based approaches at varying turbidity levels, and field tests conducted in marina environments, validate the effectiveness of the proposed approach. We have made our code open-source to facilitate reproducibility and encourage community engagement.

\end{abstract}


\section{Introduction}

Autonomous Underwater Vehicles (AUVs) play a critical role in various underwater applications, including mapping, exploration, and maintenance. These tasks often occur in turbid waters, such as sediment-laden or polluted environments commonly found in regions with high human or animal activity. Maintenance operations, in particular, require close-range interactions with the environment, relying on precise scene understanding and accurate geometry reconstruction to plan manipulator movements effectively and avoid collisions with unmodeled obstacles. Reliable scene geometry reconstruction in turbid water conditions is therefore a cornerstone for advancing the autonomy and operational safety of AUVs.

Vision-based scene reconstruction has demonstrated partial success in underwater conditions \cite{Yang2021}; these methods often depend on the detection of salient features and are highly sensitive to illumination changes. In underwater environments, light scattering caused by suspended particles results in blurring and halo effects, while wavelength-dependent absorption leads to color distortion and signal loss. Consequently, the performance of vision-based systems in scattering media is unreliable. Additionally, when reconstructions are achieved, monocular visual methods can suffer from scale ambiguity.

Sonar sensors offer a robust alternative for working in turbid and dynamic lighting conditions, as they are unaffected by light scattering and absorption. However, sonar systems have inherent limitations: profiling sonars capture narrow slices of environmental data, while imaging sonars provide 2D projections of 3D spaces, resulting in restricted vertical fields of view. Furthermore, sonar systems lack the ability to generate dense depth estimates; 
their low resolution makes them more suitable for large-scale scene perception than for detecting fine features or close-range objects critical to manipulation tasks.

Optical imaging can complement sonar information by delivering high-resolution data, while acoustic imaging ensures reliable performance in turbid waters and provides greater observation ranges. Combining these modalities can enhance the reconstruction of underwater geometry. However, the fundamentally different image generation mechanisms of sonar and cameras complicate direct feature matching, making their integration a challenging task. Overcoming these challenges holds significant potential for critical underwater applications, such as autonomous manipulation and navigation.

Recent works demonstrate the benefits of integrating optical and acoustic sensing modalities \cite{Hu2023}. However, most existing methods rely on visual cues as the primary source of information, using sonar data only to refine their results. Such approaches are unsuitable for turbid environments where salient visual features are unreliable or unavailable.

This work introduces a novel methodology for real-time scene reconstruction in highly turbid underwater environments. The approach enhances sonar-based reconstructions by incorporating visual information to expand the sonar's field of view and add detail to the reconstruction. Unlike prior methods, our algorithm uses sonar features as the foundation, without assuming the availability of reliable visual features.
To the best of our knowledge, this is the first underwater reconstruction methodology to merge sonar and monocular camera data that is specifically optimized for performance in highly turbid underwater environments.
By addressing the limitations of existing reconstruction techniques, this work aims to improve AUV performance in complex underwater scenarios, paving the way for more reliable and effective operations in turbid environments.
An open source repository for the code used in this paper is provided to facilitate the use of this approach: \textbf{\href{https://github.com/ivanacollg/sonar\_camera\_reconstruction}{https://github.com/ivanacollg/sonar\_camera\_reconstruction}}

In the following sections, we first review related work that provides inspiration for our study. Next, we define the problem to be addressed. We then detail our proposed methodology. Subsequently, we present two experiments demonstrating that our algorithm performs comparable to state-of-the-art sonar-based and vision-based reconstruction methods under varying turbidity conditions. Finally, we present a field demonstration that illustrates the scalability of our algorithm and conclude with a summary of our findings.

\vspace{-2mm}
\section{Related Works}
\vspace{-1mm}

 \textbf{Sonar reconstruction:} Sonars are widely used in underwater environments because they are impervious to lighting conditions. Wide-aperture, forward-looking multibeam imaging sonars provide an expansive Field Of View (FOV) for safe navigation and collision avoidance. Nonetheless, imaging sonars provide a flattened 2D image of an observed 3D volume. As a result, scene reconstruction while lacking elevation information is challenging. Aykin \cite{Aykin2013} proposes a single-sonar 3D reconstruction method by leveraging shadow edges. However, this work is limited to seafloor mapping. Westman \cite{Westman2019} expands this approach to include arbitrary scenes by introducing a surface reconstruction based on generative models. Both \cite{Aykin2013} and \cite{Westman2019} assume that the objects in the scene have smooth surfaces and vary monotonically in terms of distance from the sonar. A space carving method for 3D reconstruction is suggested  by Guerneve \cite{Guerneve2018}.  Park \cite{Park2023} presents a method that requires images taken at different angles by performing a 90 degree rolling motion of the sonar. However, \cite{Guerneve2018} and \cite{Park2023} require multiple views and movement around an object, which is not always possible \color{black} when working in space-limited environments or close range applications\color{black}. McConnell \cite{McConnell2020} introduces a stereo sonar approach, where sonars with orthogonal axes independently observe the same points in the environment and associate these observations. However, the resulting 3D reconstruction is limited by the small region with overlapping fields of view. McConnell \cite{McConnell2021} later proposed a Bayesian prediction model to increase this limited FOV.

\textbf{Vision-based methods:} 
Visual SLAM has been partially successful in underwater environments \cite{Joshi2019}. The available methods include indirect ones, such as ORBSLAM \cite{ORBSLAM}, which rely on feature detection and matching,
while direct methods, such as LSD SLAM \cite{Engel2014}, bypass feature detection and matching by directly utilizing pixel intensities to minimize photometric error. To improve the state estimation performance of visual SLAM strategies, inertial sensors have been incorporated. ORBSLAM3 \cite{Campos2021} is a Visual-Intertial SLAM strategy that integrates Inertial Measurement Unit (IMU) sensing.

 Some works have attempted to make visual methods more robust to underwater conditions. Islam \cite{Islam2020} attempts real-time image enhancement for color correction and deblurring. Other works \cite{Detry2018}, \cite{Fan2024} strive to make image stereo matching for 3D reconstruction easier by using an active perception approach and projecting structured light into the scene. Similarly, onboard lights are used by Roznere \cite{Roznere2023} to aid in underwater cave reconstructions. However, \cite{Detry2018, Fan2024, Roznere2023} assume that the scene conditions allow for clear or semi-clear light projections, which is not the case when working in highly turbid waters or close to the water surface, as a low level of ambient light can disturb the effect of artificial light \cite{Rahman2018}.
Furthermore, monocular optical methods present scale ambiguity.   

\textbf{Sonar-Camera approaches:} Merging sonar and camera information shows promise in overcoming information ambiguity in each sensor, elevation and depth, respectively. However, due to differences in modalities, fusing the data is challenging. Kim \cite{Kim2019} proposes to create two separate 3D models, one optical and one acoustic; the two volumetric models then correct each other and are integrated into a hybrid model. However, this method requires high computation time, thus, it does not run in real-time. Kunz \cite{Kunz2013} proposes to add sonar as a distance constraint to an optical SLAM framework. Later, \cite{Rahman2018} and \cite{Rahman2022} also avoid direct sonar to optical image matching by augmenting SLAM with acoustic constraints. The camera and sonar are combined over multiple samples. Patches based on the visual features are created and used to determine if they correspond to the features previously observed by the sonar. A multi-session SLAM method \cite{Jang2021} suggests a style-transfer approach to morph a sonar image to resemble an optical image and thus enables simple 2D-2D matching. Abu \cite{Abu2023} presents an object-based SLAM method that matches objects in a sonar image with objects in an RGB image based on geometric similarities. A semantic SLAM method \cite{Singh2024} proposes using a neural network to detect an object in an image that could provide an elevation angle for the object viewed in the imaging sonar. Although these methods have proven successful for localization, they do not perform direct sonar-to-image matching for geometry reconstruction. 

A single beam echosounder was used by Roznere \cite{Roznere2020} to find the correct depth of a monocular SLAM system by projecting 3D sonar points onto an optical image and finding a rescaling depth ratio. This approach is extended by Cardaillac \cite{Cardaillac2023} to work for an imaging sonar. Although \cite{Roznere2020} and \cite{Cardaillac2023} do perform sonar to optical image matching, they still rely on visual SLAM to provide a 3D point cloud, which would not be reliable in turbid environments.

\textbf{Machine learning approaches:} Machine learning has been leveraged to avoid tough stereo matching of underwater optical images to obtain 3D information. Some works propose using neural networks to estimate depth information directly from single monocular images \cite{Yu2023, Yang2024}. The depth information in \cite{Liu2023} was estimated using a deep neural network that takes an optical image and a single-beam ecosounder as input. The use of Gaussian splatting for 3D reconstruction using optical image and imaging sonar was presented by Qu \cite{Qu2024}. AONeuS, a neural rendering framework for acoustic-optical sensor fusion, was introduced by Qadri \cite{Qadri2024}. Although machine learning-based approaches show promising results, they require training, which is challenging as underwater data is limited and underwater conditions vary dramatically, negatively impacting performance. 

\textbf{Turbidity-conscious approaches:} 
Babaee \cite{Babaee2013} aims to make sonar to optical image matching more robust to turbidity by using contours. Limiting the area of feature match search by detecting a bounded area of interest was proposed by Spears \cite{Spears2014}, however, this still requires detecting point features in an optical image, which is not possible at times when images in turbid waters are highly blurred. Yevgeni \cite{Yevgeni2024} avoids having to detect point features in optical images by detecting relevant areas of interest instead and matching them directly to sonar depth values. Our work relates the closest to the works in this last category and presents a sonar-camera fusion method for reconstructing scenes in highly turbid waters. 

\vspace{-2mm}
\section{Problem Formulation}
\begin{figure}[tb]
    \centering
    \begin{subfigure}{0.4 \columnwidth}
        \centering
        \includegraphics[width=\linewidth]{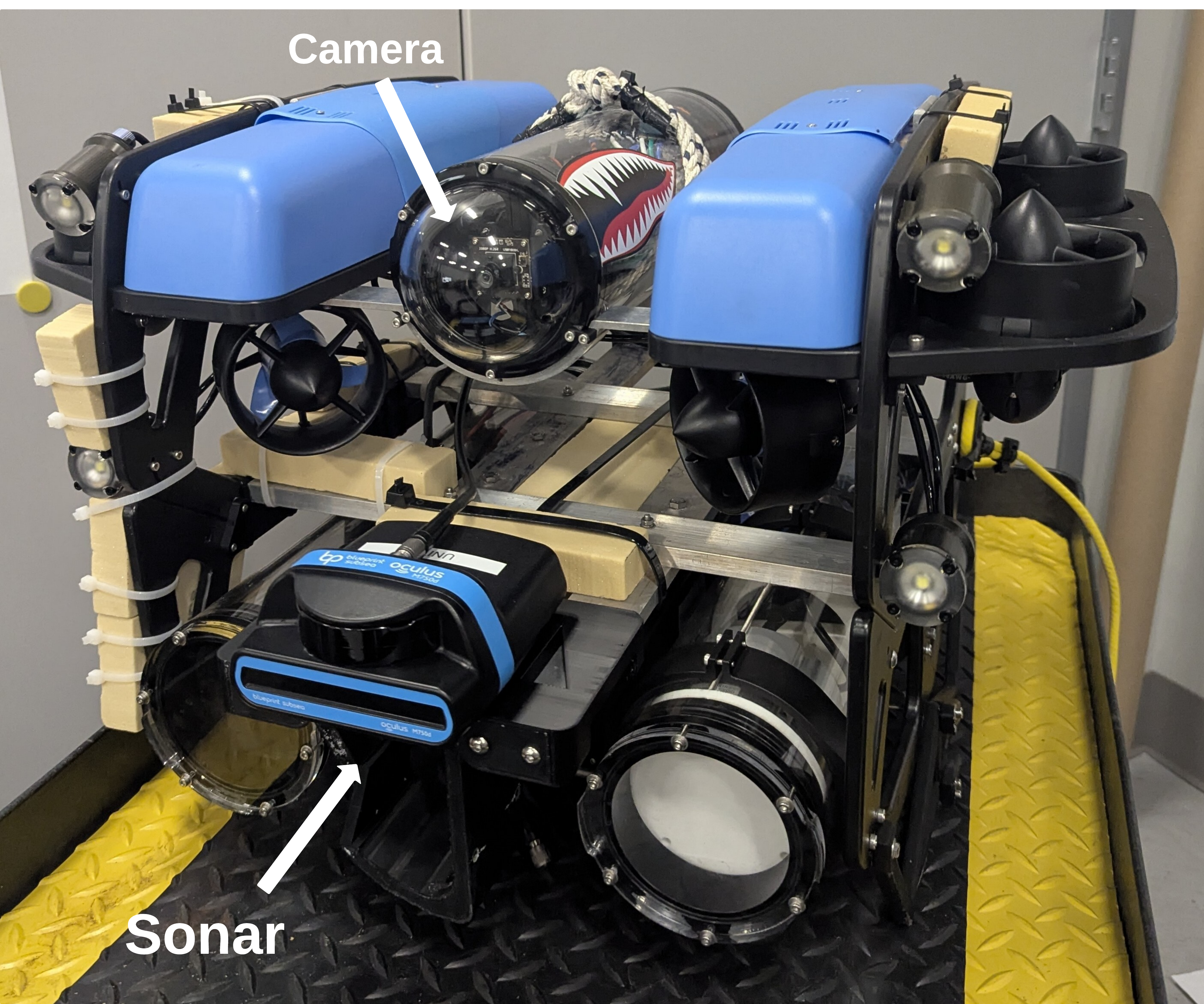} 
        \caption{BlueROV2 platform.}
        \label{fig:ROV}
    \end{subfigure}
    \begin{subfigure}{0.45\columnwidth}
        \centering
        \includegraphics[width=\linewidth]{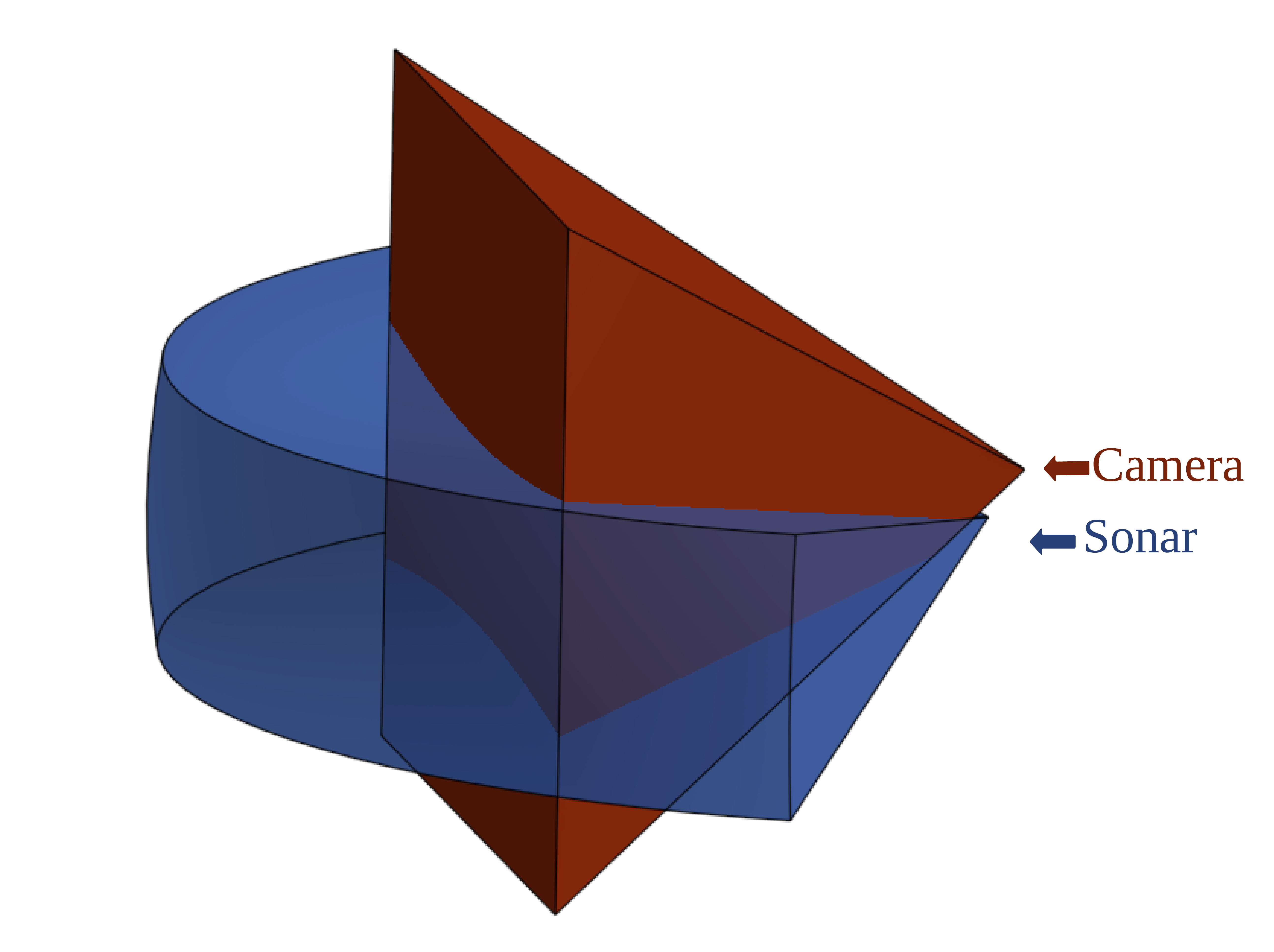} 
        \caption{Sonar and camera fields of view.}
        \label{fig:FOV}
    \end{subfigure}
    \caption{\textbf{Fields of view} corresponding to the hardware
arrangement in (a) is shown in (b); the red swath is from the
camera and the blue is from the sonar.}
    \label{fig:FOVandROV} \vspace{-5mm}
\end{figure}

This work considers reconstructing an environment using an imaging sonar and a monocular camera. We assume that a robot is equipped with a forward-looking imaging sonar and a forward-looking camera, and the sensors are mounted such that their fields of view overlap, as shown in Fig. \ref{fig:FOVandROV}. This overlap allows some areas to be observed simultaneously by both sensors. This implies that with proper calibration, a point observed by both sensors, corresponds to the same location.


\vspace{-2mm}
\subsection{Camera Model}
Environments are represented as a collection of points $\mathbf{P} \in \mathbb{R}^3$. Transforming a 3D point from the sonar reference frame $\mathbf{P}_s$ to the camera reference frame $\mathbf{P}_c$ can be done using the extrinsic parameters $\mathbf{R}_{s \to c}$ and $\mathbf{t}_{s\to c}$, where $\mathbf{R}_{s \to c}$is the rotation matrix and $\mathbf{t}_{s\to c}$ is the translation vector between the sonar and camera frames.
\begin{equation}
    \mathbf{P}_c = 
    \begin{pmatrix}
        x_c \\
        y_c \\
        z_c 
    \end{pmatrix}
    =\mathbf{R}_{s \to c}\mathbf{P}_s+\mathbf{t}_{s\to c}
\end{equation}
Using the pinhole camera model, a 3D point in the camera reference frame $\mathbf{P}_c$ can be converted into a 2D point $\mathbf{p}_c$ using the intrinsic camera matrix $\mathbf{K}$:
\begin{equation}
    \mathbf{p}_c = \begin{pmatrix}
        u \\
        v \\
        1 
    \end{pmatrix} = \frac{\mathbf{P}_c}{z_c}\mathbf{K},
     \text{with } \mathbf{K} =
        \begin{pmatrix}
        fx & 0 & c_u\\
        0 & f_y & c_v\\
        0 & 0 & 1
    \end{pmatrix}.
    \label{eq:3Dto2D}
\end{equation}
$(f_x, fy)$ define the camera focal length, and $(c_u,c_v)$ define the camera optical center.

\vspace{-2mm}
\subsection{Sonar Model}
\begin{figure}[tb]
\centerline{\includegraphics[width=0.4\textwidth]{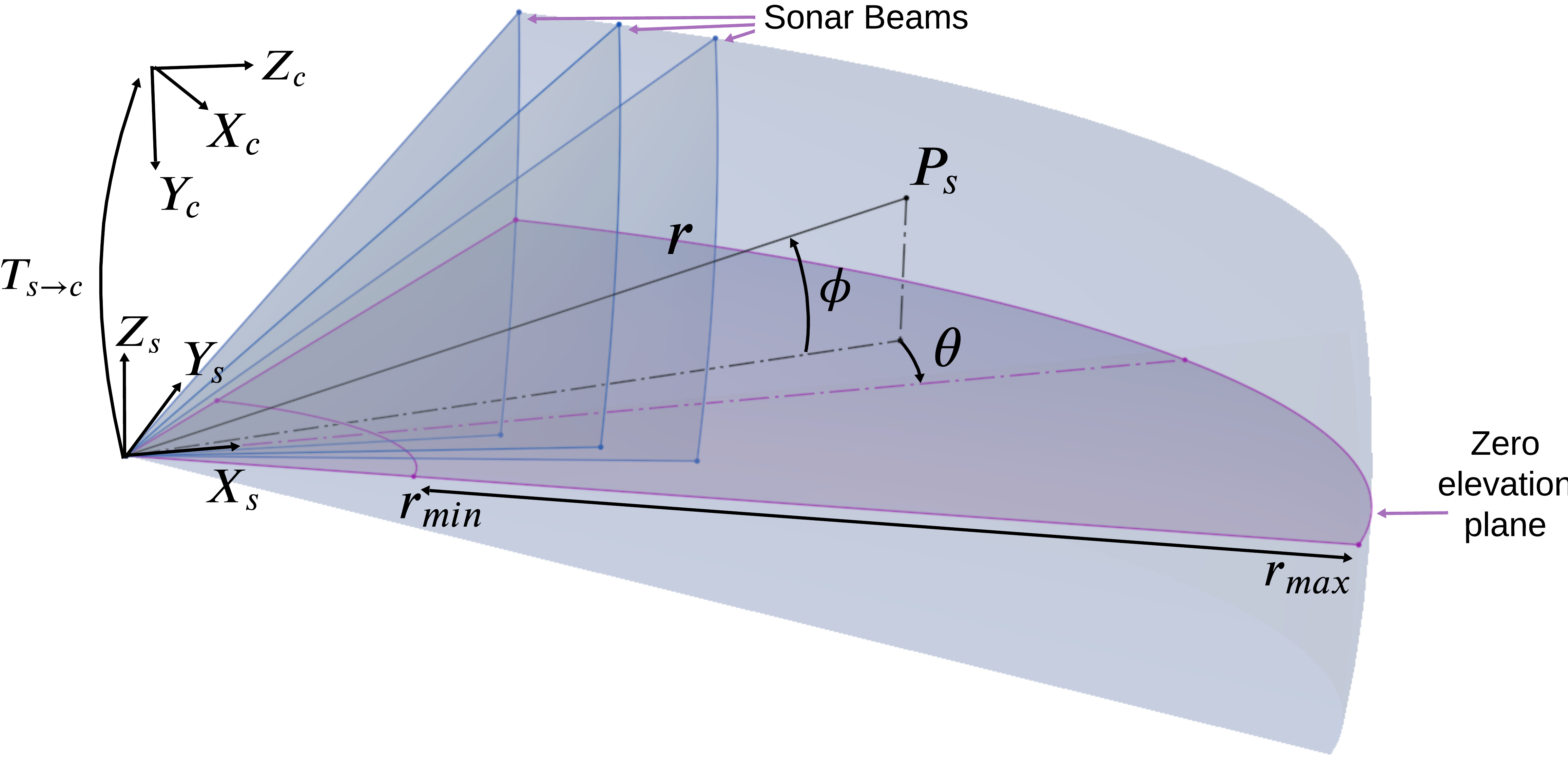}}
\caption{\textbf{Forward looking imaging sonar model.} The point $\mathbf{P}_s$ can be represented by $[r, \theta, \phi]^T$ in a spherical coordinate frame. The range $r$ and the bearing angle $\theta$ of $\mathbf{P}_s$ are measured, while the elevation angle $\phi$ is not captured in the resulting 2D sonar image. The 3D transformation $\mathbf{T}_{s \to c}$ from sonar to camera frame is also depicted. }
\label{fig:Sonar} \vspace{-4mm}
\end{figure}

The multibeam forward looking sonar transmits acoustic signals and measures their reflected intensities. The imaging sonar senses a 3D volume depicted in Fig. \ref{fig:Sonar}. Points in 3D space are denoted in spherical coordinates $[r, \theta, \phi]^T$, corresponding to range $r \in \mathbb{R}_+$, bearing $\theta \in \Theta$ and elevation angle $\phi \in \Phi$, where $\Theta, \Phi \subseteq [-\pi, \pi)$. Each acoustic ping return represents a beam with elevation values $\phi \in [\phi_{max}, \phi_{min}]$, meaning the sonar returns a 2D intensity array without elevation information. 
The transformation of a 3D point from spherical to Cartesian coordinates with respect to the sonar reference frame, is described by \vspace{-2mm}
\begin{equation}
    \mathbf{P}_s = 
    \begin{pmatrix}
        x_s\\
        y_s\\
        z_s
    \end{pmatrix}
    =r
    \begin{pmatrix}
        \cos\phi \cos\theta\\
        \cos\phi \sin\theta\\
        \sin\phi
    \end{pmatrix}.
    \label{eq:sonar3D}
\end{equation}

\vspace{-3mm}
\subsection{Camera-Sonar Correspondence Problem}
As shown in Eq. (\ref{eq:sonar3D}), a precise elevation angle $\phi$ is needed to obtain the 3D locations of the sonar data. Likewise, distance information $z_c$ is necessary for precise projection of 3D points onto the camera image plane (Eq. (\ref{eq:3Dto2D})). Because the sonar produces distance information and the optical camera offers elevation information, with both sensors contributing bearing data, their complementary information can potentially be correlated to obtain full 3D data. Nevertheless, performing opti-acoustic integration is a challenging task due to their distinct image generation mechanisms which do not allow simple feature matching techniques to solve the transformation problem.

\vspace{-4mm}
\section{Proposed Approach} 
\begin{figure*}
\centerline{\includegraphics[width=0.99\linewidth]{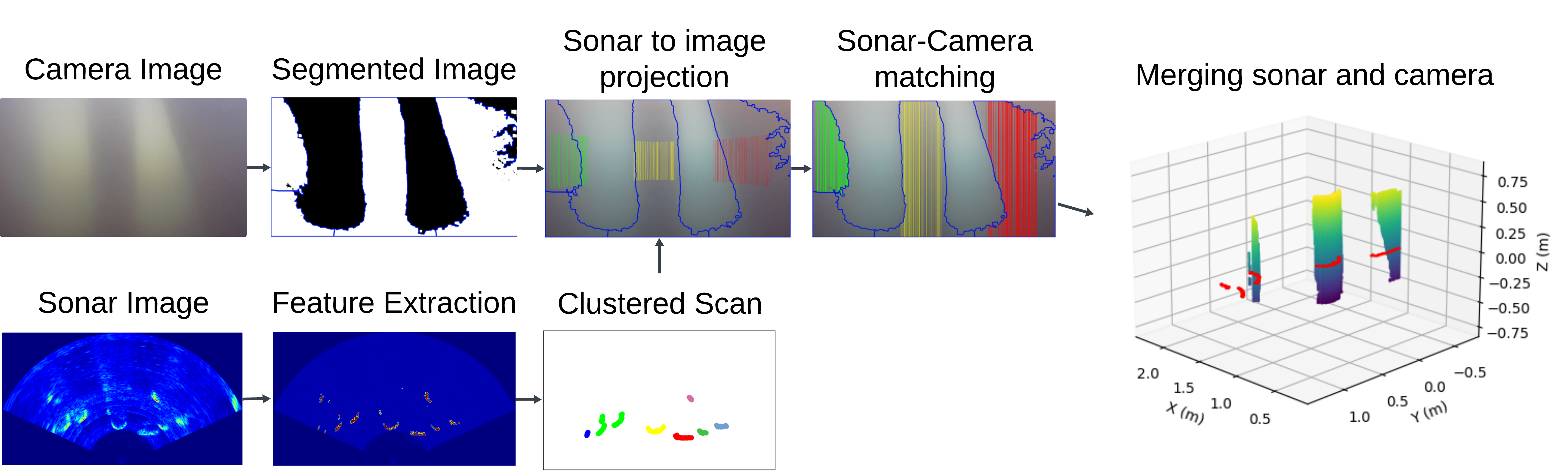}} 
\caption{\textbf{System Architecture:} An overview of the proposed pipeline.} \label{fig:method} \vspace{-5mm}
\end{figure*}

In this section, we describe our proposed framework for identifying correspondences between monocular optical images and sonar images. An overview of the system can be observed in Fig. \ref{fig:method}. The primary goal of this pipeline is to associate range measurements from sonar with elevation measurements from optical images, leveraging their overlapping but complementary vantage points, each lacking a dimension in their respective coordinate frames. The algorithm outputs a set of fully defined points in 3D Euclidean space.

\subsection{Optical Image Segmentation}

Underwater images often suffer from poor contrast, color attenuation, and non-uniform lighting. Consequently, point features cannot be reliably detected under these conditions. Instead of detecting point features in a noisy, low-contrast underwater image, we propose detecting regions of interest corresponding to objects in the scene.

We first apply a Gaussian filter to reduce noise in the image, followed by an adaptive threshold filter to locally enhance regions of the image under non-uniform lighting conditions. The image is then segmented into regions of interest using the watershed algorithm, which outputs independent regions $R$. To exclude noise and particles, we only retain regions larger than a specific threshold of $T$ pixels. We express this as: 
\begin{equation} I_{\text{segmented}} = \bigcup_{i=1}^k R_i, \quad \text{such that} \quad |R_i| \geq T,  
\end{equation} 
where $k$ is the number of regions in the image $I$. The result is a binary image highlighting only the regions of interest.

\vspace{-2mm}
\subsection{Sonar Processing}
The 2D intensity array returned by the sonar can be arranged into a polar image format with coordinates representing the range $r$ and bearing $\theta$ of each return. However, not all pixels in the acoustic image contain relevant intensity information; noise and second returns are common. Therefore, preprocessing is necessary to filter noise and erroneous measurements.

We apply the SOCA-CFAR filter \cite{El-Darymli2013}, a variant of the Constant False Alarm Rate (CFAR) technique \cite{Richards2005}, to detect features. This filter effectively removes second returns that are often present in sonar images \cite{McConnell2020}.

After identifying features in the sonar image, we retain only the closest return at each bearing angle, assuming that the objects visible in the optical image are the foreground 
\color{black}object surfaces \color{black} closest to the robot. The remaining returns are clustered using Density-Based Spatial Clustering of Applications with Noise (DBSCAN) \cite{DBSCAN}, which does not require prior knowledge of the scene. The clustering results in a set
$\mathcal{C} = \{\mathcal{C}_1, \mathcal{C}_2, \dots, \mathcal{C}_m\}$,
where $m$ is the total number of clusters. Each cluster $\mathcal{C}_i=\{\mathbf{p}_{s1},\mathbf{p}_{s2},…,\mathbf{p}_{sn}\}$ contains $n$ 2D points defined by polar coordinates $\mathbf{p}_{s}=(r,\theta)$.

\vspace{-2mm}
\subsection{Sonar Beam to Image Projection}

Due to elevation ambiguity in the sonar data, each sonar return potentially corresponds to multiple elevations. We project the filtered sonar data to all possible elevation angles $\phi \in [\phi_{max}, \phi_{min}]$. Thus, each sonar return $\mathbf{p}_{s}$ is mapped to a set of possible 3D points $\mathcal{S}_j$: 
\begin{equation}
    \mathcal{S}_j = \{\mathbf{P}_s\}=r
    \begin{pmatrix}
        \cos\phi_i \cos\theta\\
        \cos\phi_i \sin\theta\\
        \sin\phi_i
    \end{pmatrix}
    \quad \text{where} \quad \phi_i \in [\phi_{max}, \phi_{min}].
\end{equation}
Each set $\mathcal{S}_j$ is then projected onto the camera image using the camera intrinsic matrix $\mathbf{K}$ and the transformation matrix $\mathbf{T}_{s \to c}$, which maps sonar coordinates to the camera frame: 
\begin{equation}
    \mathbf{p} = 
    \begin{pmatrix}
        u \\
        v \\
        w 
    \end{pmatrix}/w
    = \mathbf{K} \mathbf{T}_{s \to c}
    \begin{pmatrix}
        \mathbf{P}_s \\
         1 \\ 
    \end{pmatrix}.
\end{equation}
Each set of points $\mathcal{S}_j$ from a sonar beam can be perceived as a vertical line in the camera image. \color{black}Sonar beams appear as vertical lines in the image due to the fixed extrinsic calibration between the sonar and camera.\color{black}

\subsection{Sonar-Camera Matching}

After projecting sonar returns onto the camera image, each segmented image region $R$ is matched to a sonar cluster $\mathcal{C}$ if they overlap, satisfying $
\mathcal{C} \cap R \neq \emptyset$. 
If $R$ overlaps with multiple clusters, 
it is matched with the closest one \color{black}to prioritize closer objects that are more likely to appear in the visual image\color{black}: 
\begin{equation} 
\mathcal{C}^* = \arg\min_{\mathcal{C}_i \in {\mathcal{C} \cap R}} d(\mathcal{C}_i, R), \end{equation} 
where $ d(\mathcal{C}_i, R)$ represents the Euclidean distance between the robot and the centroid of cluster $\mathcal{C}_i$ within region $R$. For each overlapping region $\mathcal{C}^* \cap R$, we vertically expand the sonar FOV to match the shape of the region in the optical image.  An example of the expanded points can be seen in Fig. \ref{fig:method}. 

In order to project the expanded 2D points back to 3D space, each point needs a distance $z_c$ value. The distance values $z_c$ of the sonar beam points $S_j$ vary slightly depending on the elevation angle $\phi_i$ to which the point corresponds. In order to find the most accurate distance value of each pixel in the camera image we compute the mean distance $\bar{z_{c}}$ of each column of points within the overlap area $\mathcal{C}^* \cap R$. We then use $\bar{z_c}$ as the distance value for all points in that image column within the region $R$. The expanded 2D points can then be projected back to 3D space using: 
\begin{equation}
    \mathbf{P}_c =
     \mathbf{\bar{z}}_c \mathbf{K}^{-1} \mathbf{p}.
\end{equation}

Thus, the distance is derived from the original sonar return, while the elevation is determined by the optical image. This process yields a set of fully defined 3D points in the camera coordinate frame. 
We note that this depends on the assumption that for each bearing angle (i.e., sonar beam), all corresponding elevation angles in the optical image are represented by a single sonar-derived range measurement. Nonetheless, due to our deployment of the proposed method at close range to the objects of interest, it successfully reconstructs objects with varying cross-sections and angular orientations. 
\color{black}

\vspace{-2mm}
\section{Experiments and Results}
To validate the proposed camera-sonar fusion framework, we present 
experiments showing the performance of the framework under a variety of turbidity conditions. 

\vspace{-2mm}
\subsection{Experimental Setup}
The platform used for data collection is shown in Fig. \ref{fig:ROV}; it is a heavy configuration BlueROV2 underwater robot customized with a sensor payload that includes: a VectorNav VN-100 IMU running at 200 Hz, a KVH DSP-1760 Fiber Optic Gyroscope (FOG) sensing at 250 Hz, a Bar30 pressure sensor at 5 Hz, a Rowe SeaPilot Doppler Velocity Log (DVL) at 5 Hz, a camera logging at 5 Hz, and a sonar running at 5 Hz. The camera used was a Low-Light HD Sony Exmor IMX322 / IMX323; this camera uses a large physical pixel size which allows for maximum light sensitivity. The sonar used was an Oculus M750d forward-looking multibeam imaging sonar, which was operated at its low-frequency, wide aperture setting of 750kHz. In this mode, this sensor has a vertical aperture of 20$^\circ$ and a horizontal aperture of 130$^\circ$. The sonar and camera were mounted as shown in Fig. \ref{fig:FOVandROV}. \color{black} Additionally, an M1200d sonar was mounted in stereo with our M750d sonar, orthogonal to it; its data is only used for comparison. \color{black}

The robot carries an on-board Pixhawk, Raspberry Pi
and Jetson Nano for control and computation. We use ROS to operate the vehicle and log data from a topside computer. All algorithms are applied to real-time playback of our data using a computer equipped with a Titan RTX GPU and Intel i9 3.6GHz CPU (intended to represent hardware that could be fully embedded aboard an 
AUV).
The min. region $R$ size for all experiments was empirically tuned to $T=334$ pixels.

We compare our approach against four other strategies: Stereo Sonar \cite{McConnell2020}, ORBSLAM3 \cite{Campos2021}, LSD SLAM \cite{Engel2014}, and COLMAP \cite{schoenberger2016}. The methods were chosen to provide a range of suitable sparse \cite{Campos2021} and dense \cite{McConnell2020, Engel2014, schoenberger2016} representations, encompassing 
SLAM \cite{Campos2021, Engel2014} and SfM \cite{schoenberger2016} strategies, as well as vision-based \cite{Campos2021, Engel2014, schoenberger2016}, and sonar-based \cite{McConnell2020} sensing methods. Although COLMAP does not run in real-time like the other methods, it has been used for prior comparisons with underwater reconstruction methods \cite{Wang2023}. \color{black} We did not include AONeuS in our evaluation; it is not designed for real-time use nor generalizable across different scenes.\color{black}


Although the Stereo Sonar algorithm and ours can run using the observations from a single robot pose, other state-of-the-art reconstruction methods require movement to perform feature matching between images and obtain a reconstruction. Therefore, to compare our strategy with others, we estimate the ROV's trajectory simply by integrating in time the DVL's velocities, the FOG and IMU's angular rates, and by using pressure sensor readings for depth. The point cloud output of our algorithm is then aggregated together at each time step. 


To compute coverage, the final point clouds are voxelized by iterating over all the points and placing them in their respective discrete bins. If a voxel contains one or more points, it is counted; otherwise, it is not. Our study uses voxel count to quantify coverage, avoiding the over-counting of redundant information contained in the point cloud. We analyze the coverage of the resulting point clouds by assessing the total number of filled voxels using a 1cm grid cell resolution.

\vspace{-1mm}
\subsection{Turbidity emulation}
\vspace{-1mm}
In order to evaluate performance in different water turbidity conditions, the underwater visual data of our indoor tank experiments was modified to resemble different levels of turbidity. The turbidity effects were simulated following the underwater image synthesis method proposed in \cite{Raihan2022}. This method is based on the Image Formation Model (IFM)  
\begin{equation}
I^c(x) = J^c(x)t^c(x)+(1-t^c(x))B^c ;c\in{R,G,B},
\end{equation}
where $I^c$ is the resulting modified image. $B^c$ represents the scattering of background light as it travels toward the camera. The statistical average background light values per color channel of $B_r = 0.6240$, $B_g = 0.805$, and $B_b = 0.7651$ were used. The transmission map $t^c(x)$ is estimated by
\vspace{-1mm}
\begin{equation}
t^c=e^{-\beta^cd(x)};c\in{R,G,B}.
\end{equation}
\vspace{-1mm}
A static water depth $d(x)$ value of 1m was chosen for the purpose of our trials - the robot was operated at this depth in our tank, and this is also the approximate distance between the robot and the observed structures. The water types are broadly classified as oceanic and coastal waters. The oceanic water types include type-I, type-IA, type-IB, type-II, and type-III subcategories, and the coastal water types include type-1C, type-3C, type-5C, type-7C, and type-9C, from clear to turbid. The water attenuation coefficients $\beta^c$ were chosen according to the Jerlov water types \cite{Jerlov1947} shown in Table \ref{tab:attenuation_coefficients}. \vspace{-2mm}
\begin{table}[h]
    \centering
    \begin{tabular}{|l|c|c|c|}
    \hline
    \textbf{Water type} & \textbf{Red channel} & \textbf{Blue channel} & \textbf{Green channel} \\ \hline
    Type-I    & 0.85 & 0.96 & 0.98 \\ \hline
    Type-5C   & 0.67 & 0.73 & 0.67 \\ \hline
    Type-7C   & 0.62 & 0.61 & 0.50 \\ \hline
    Type-9C   & 0.55 & 0.46 & 0.29 \\ \hline
    \end{tabular}
    \caption{Water attenuation coefficients chosen based on Jerlov  types \cite{Jerlov1947}.} \vspace{-3mm}
    \label{tab:attenuation_coefficients}
\end{table}

\begin{figure*}[tbh]
    \centering
    \begin{subfigure}{0.24\linewidth}
        \centering
        \includegraphics[width=\linewidth]{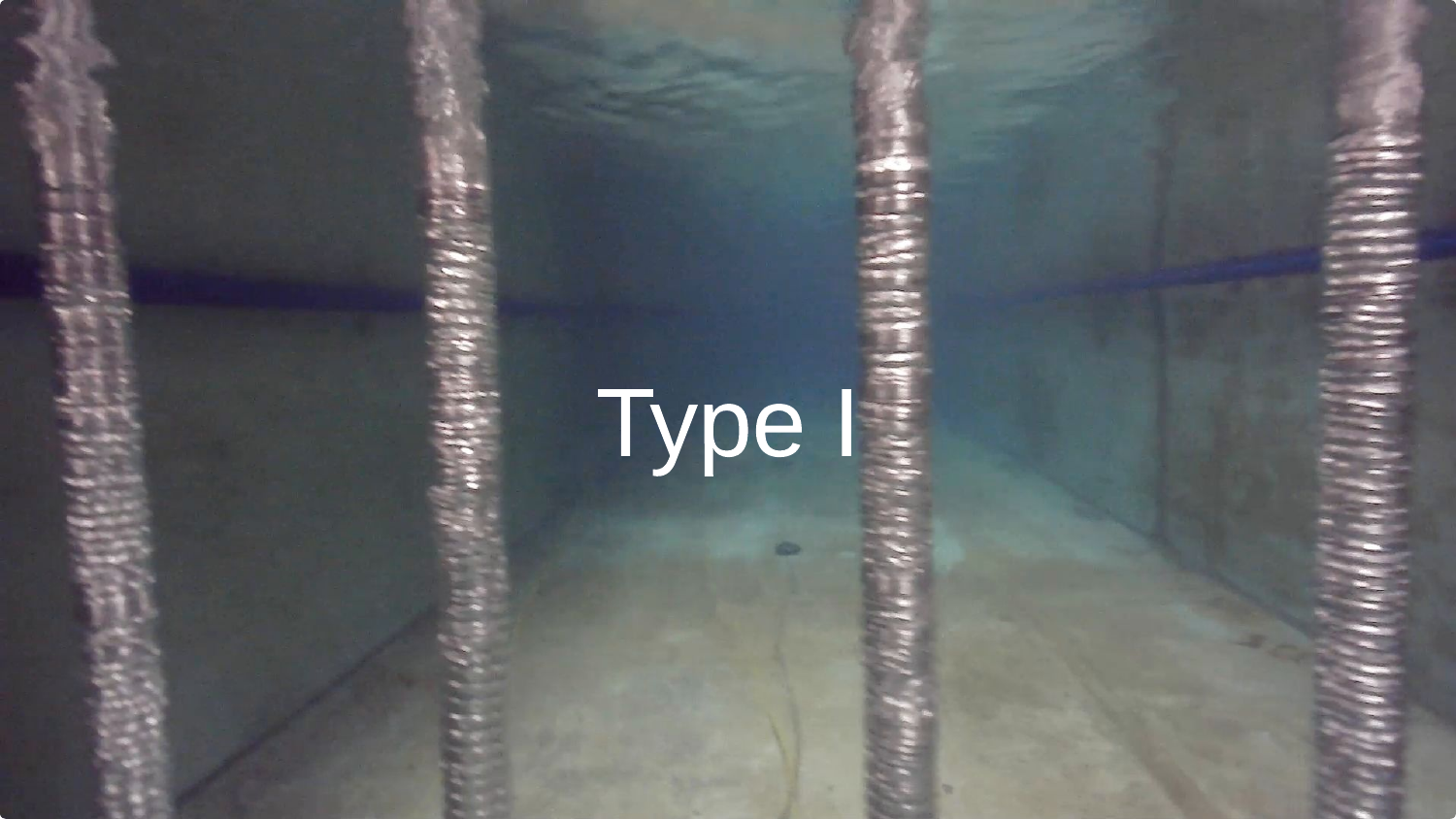} 
    \end{subfigure}
    \begin{subfigure}{0.24\linewidth}
        \centering
        \includegraphics[width=\linewidth]{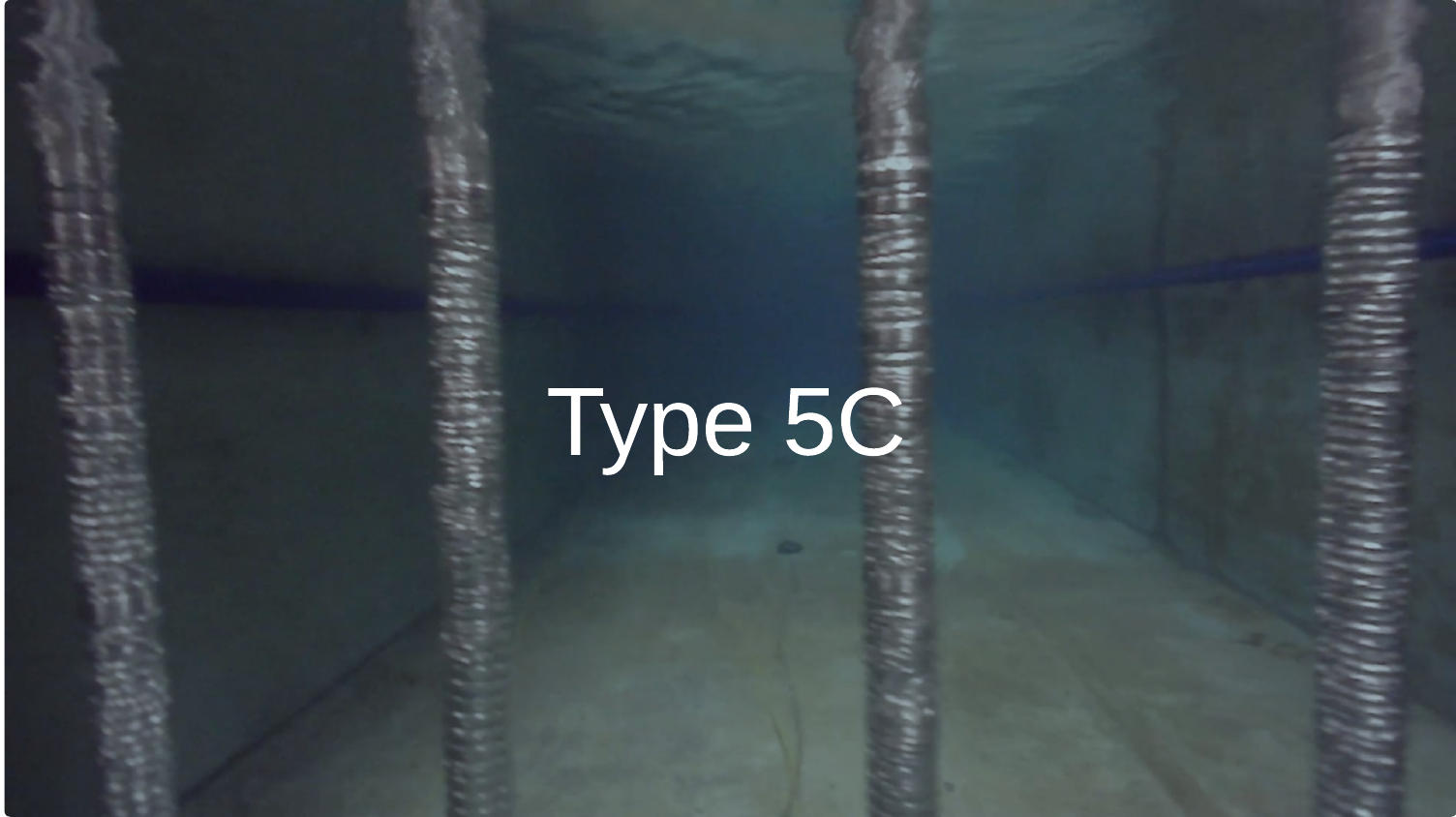} 
    \end{subfigure}
    \vspace{0.1cm}
    \begin{subfigure}{0.24\linewidth}
        \centering
        \includegraphics[width=\linewidth]{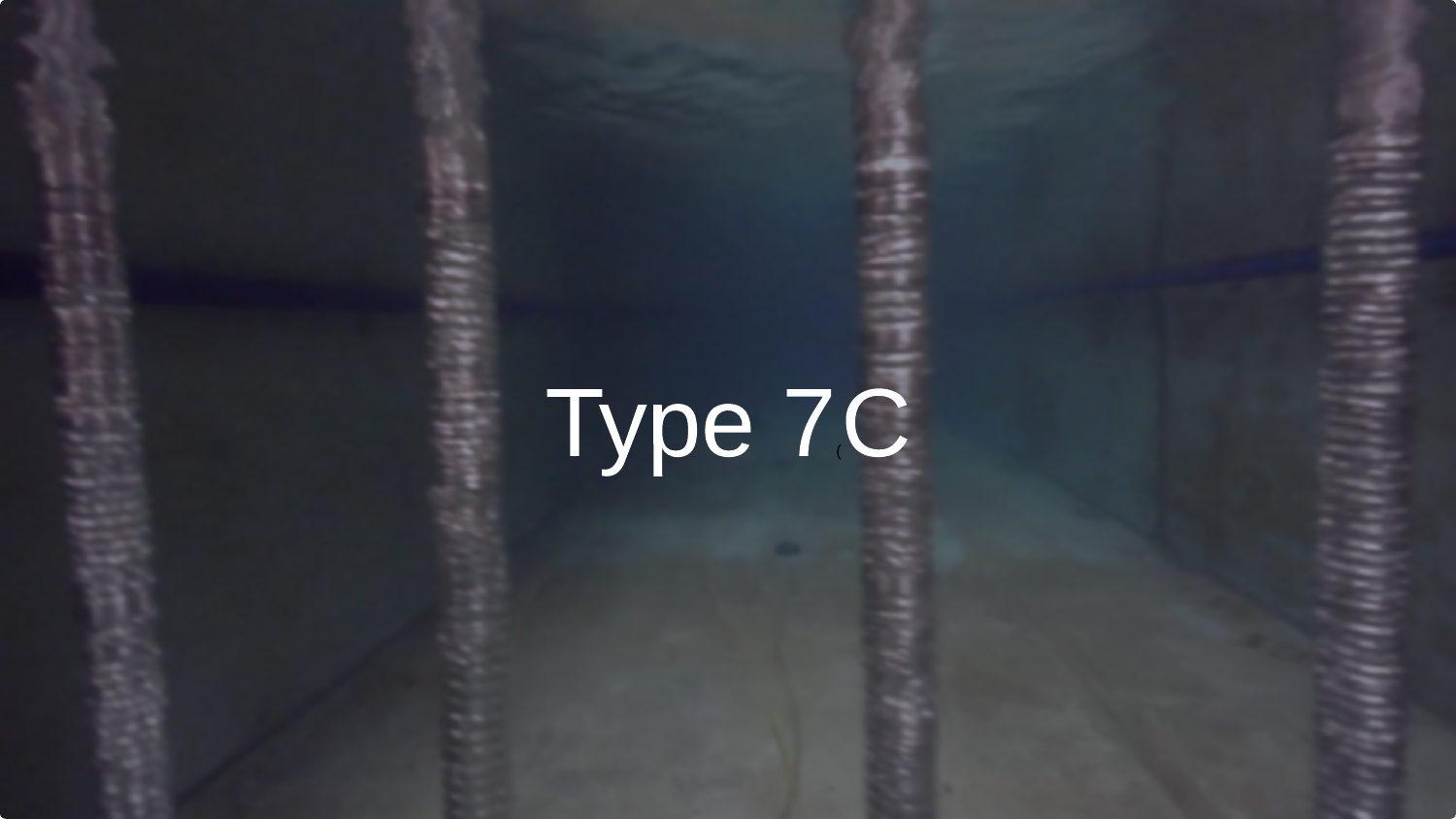} 
    \end{subfigure}
    \begin{subfigure}{0.24\linewidth}
        \centering
        \includegraphics[width=\linewidth]{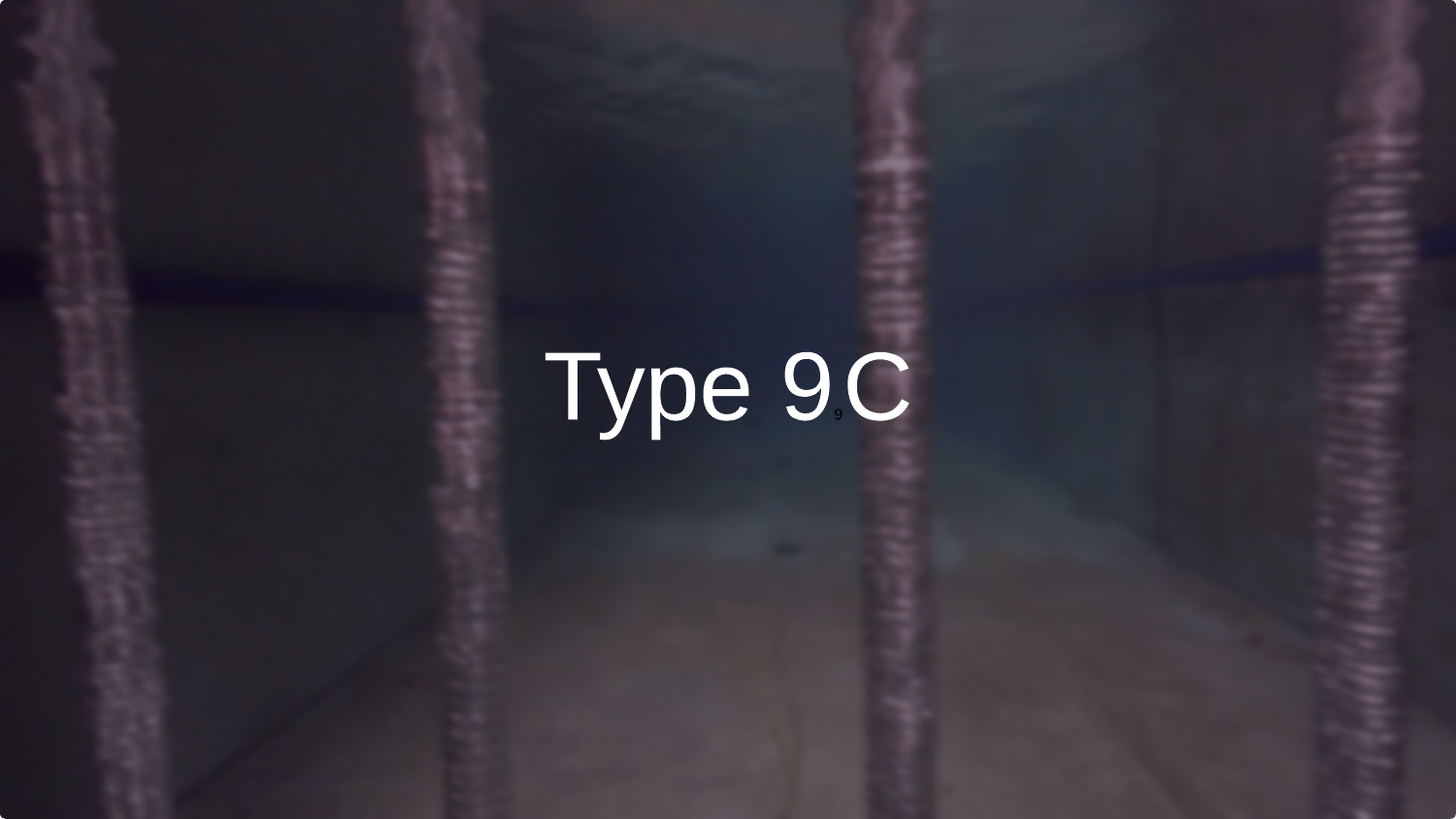} 
    \end{subfigure}
    \begin{subfigure}{0.23\linewidth}
        \centering
        \includegraphics[width=\linewidth]{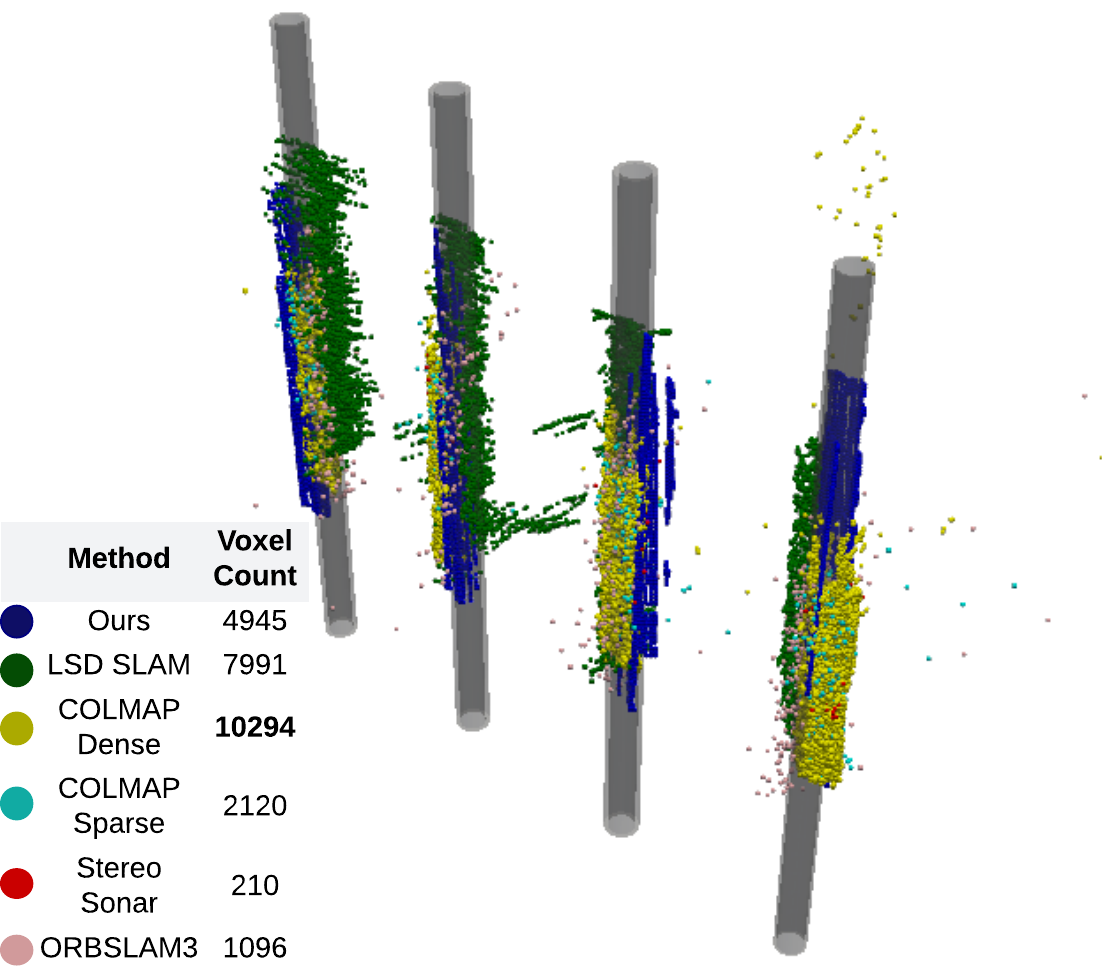} 
    \end{subfigure}
      \hfill
    \begin{subfigure}{0.24\linewidth}
        \centering
        \includegraphics[width=\linewidth]{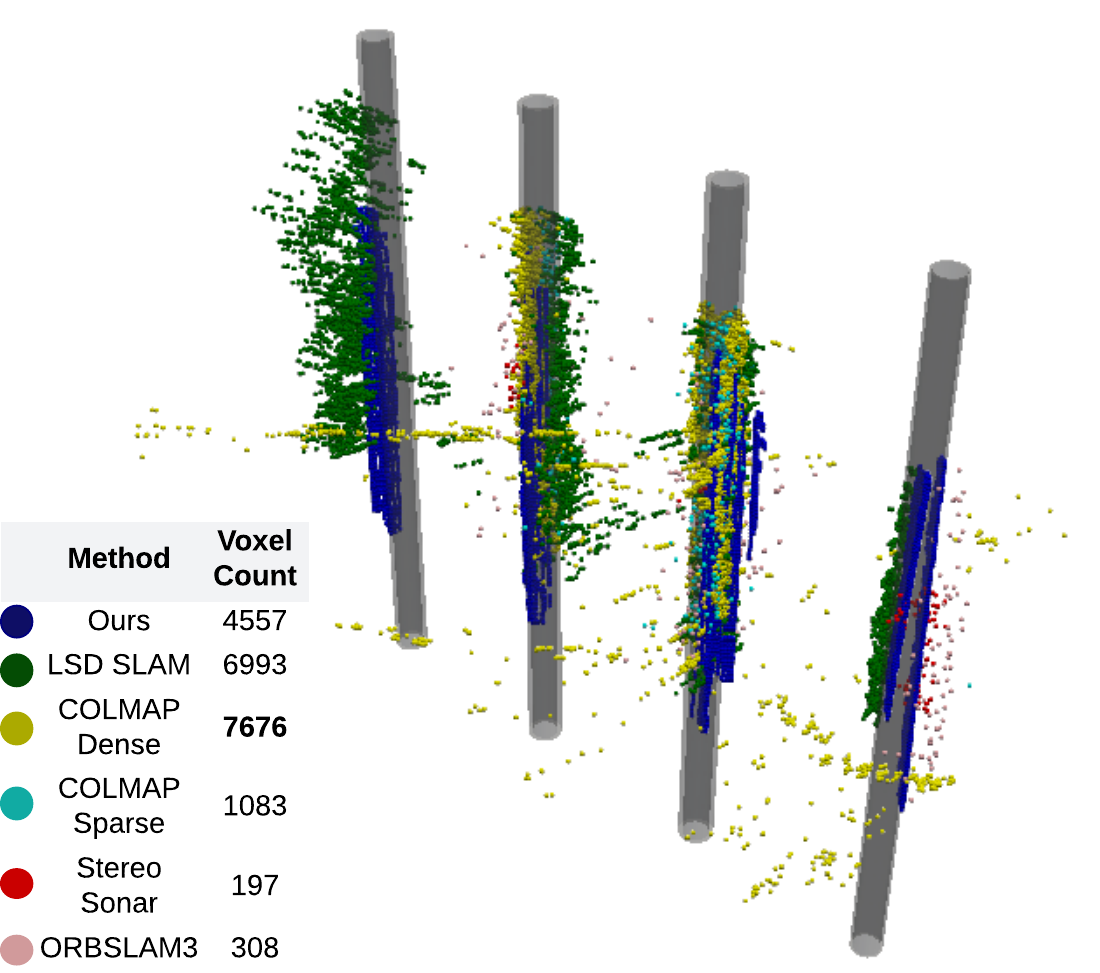} 
    \end{subfigure}
      \hfill
    \begin{subfigure}{0.24\linewidth}
        \centering
        \includegraphics[width=\linewidth]{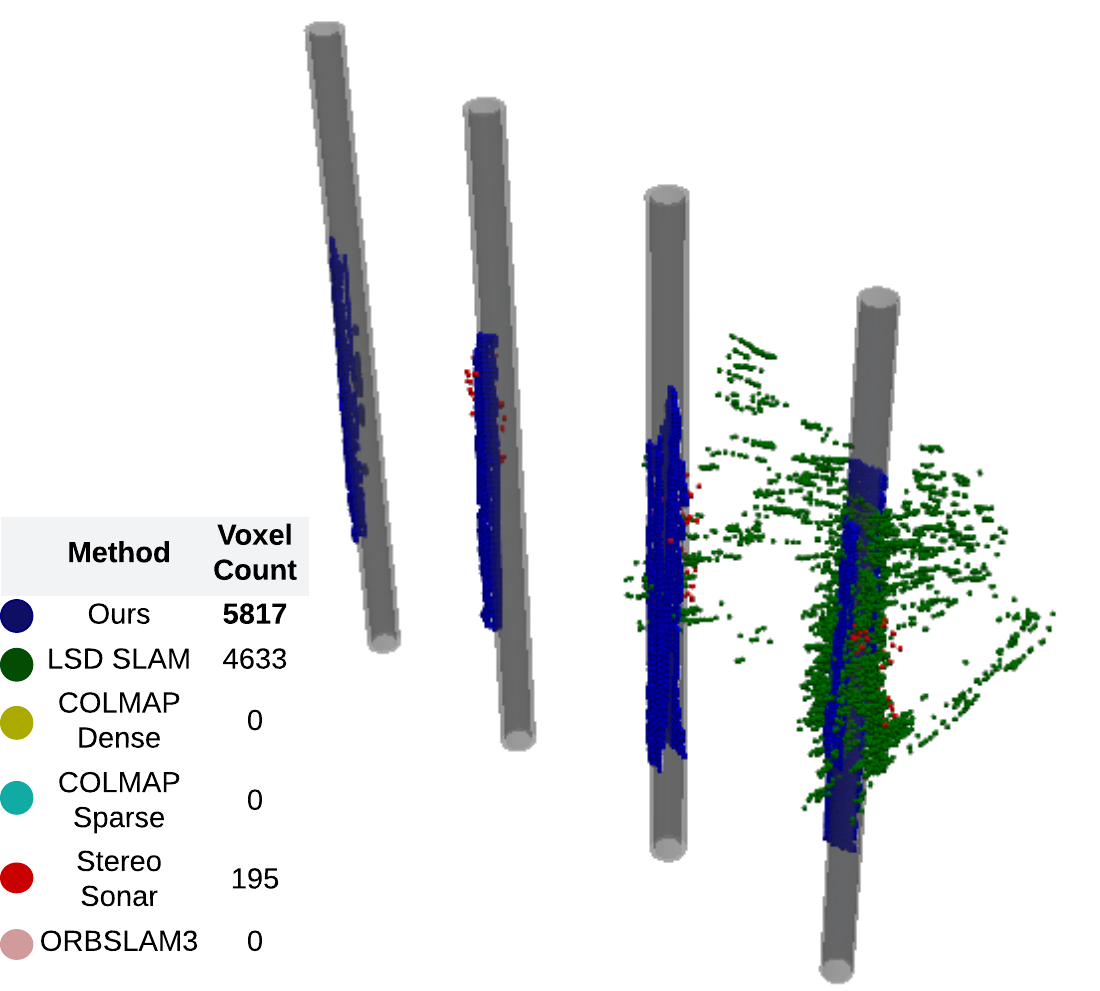} 
    \end{subfigure}
      \hfill
    \begin{subfigure}{0.24\linewidth}
        \centering
        \includegraphics[width=\linewidth]{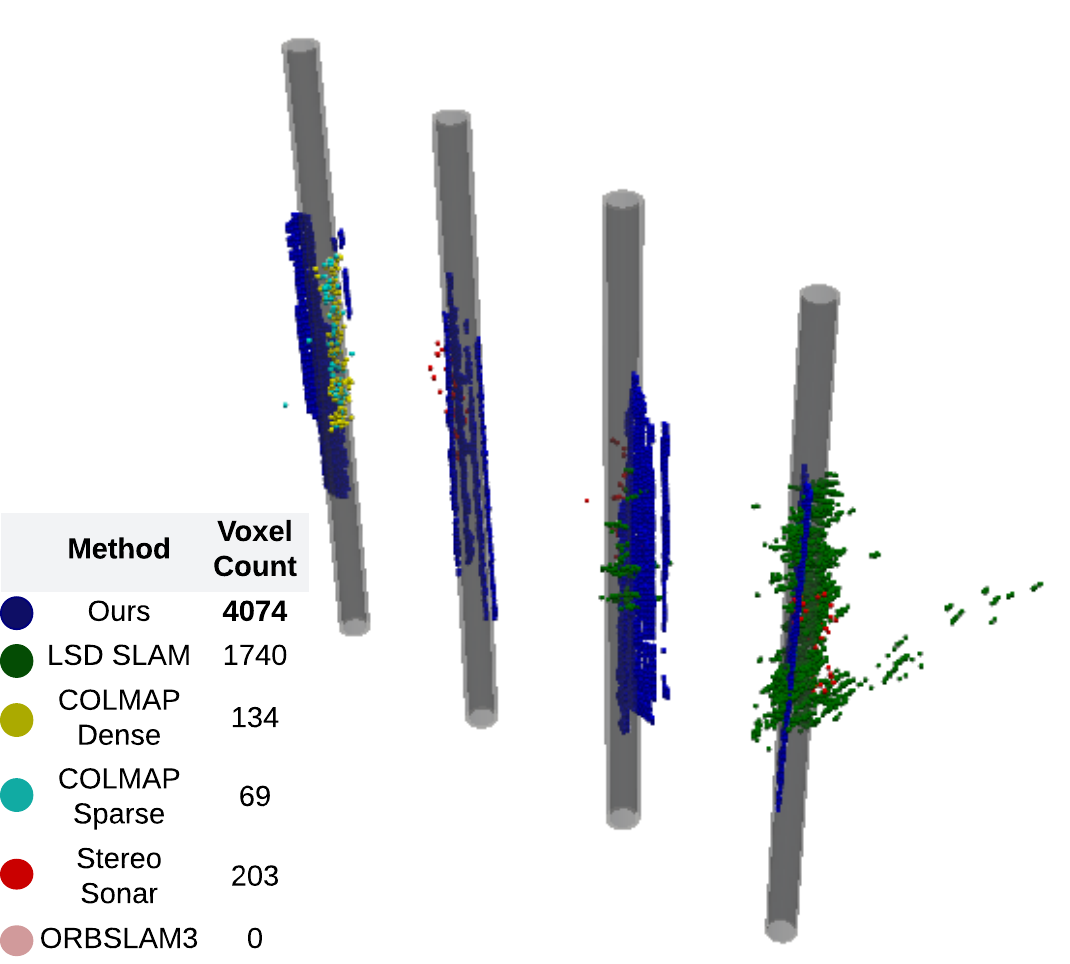} 
    \end{subfigure}

    \begin{subfigure}{0.24\linewidth}
        \centering
        \includegraphics[width=\linewidth]{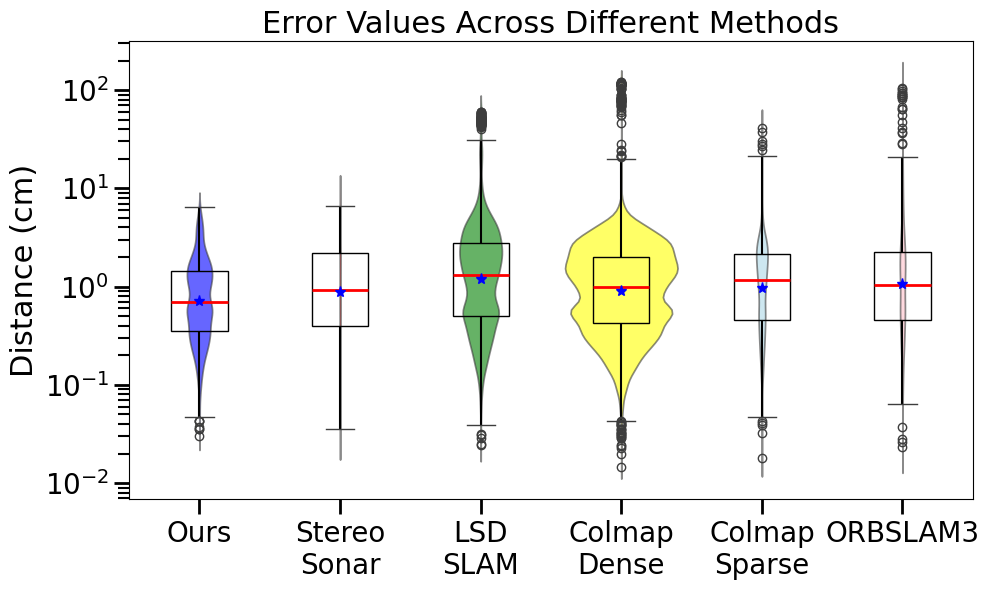} 
    \end{subfigure}
    \begin{subfigure}{0.24\linewidth}
        \centering
        \includegraphics[width=\linewidth]{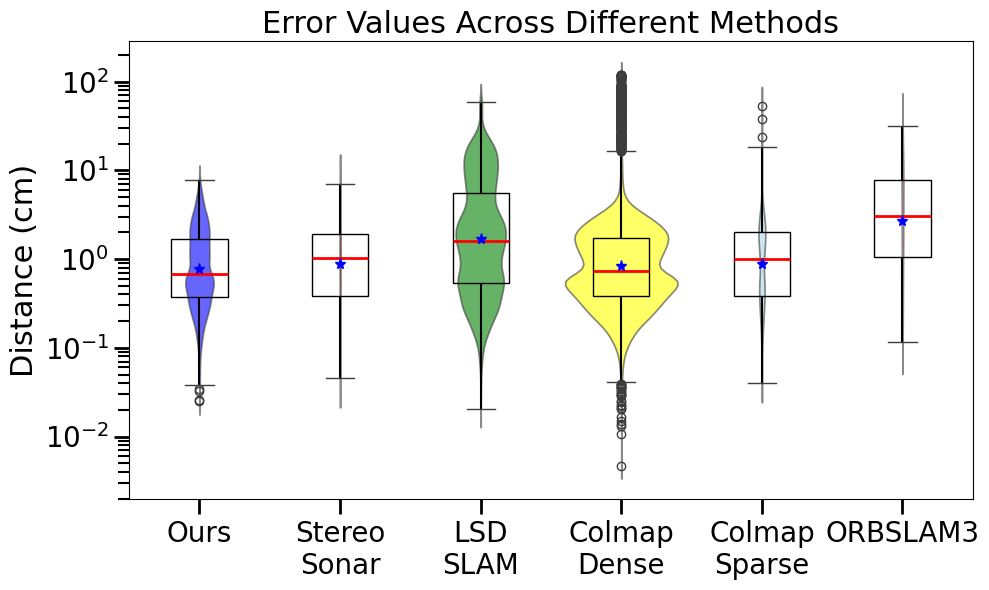} 
    \end{subfigure}
    \begin{subfigure}{0.24\linewidth}
        \centering
        \includegraphics[width=\linewidth]{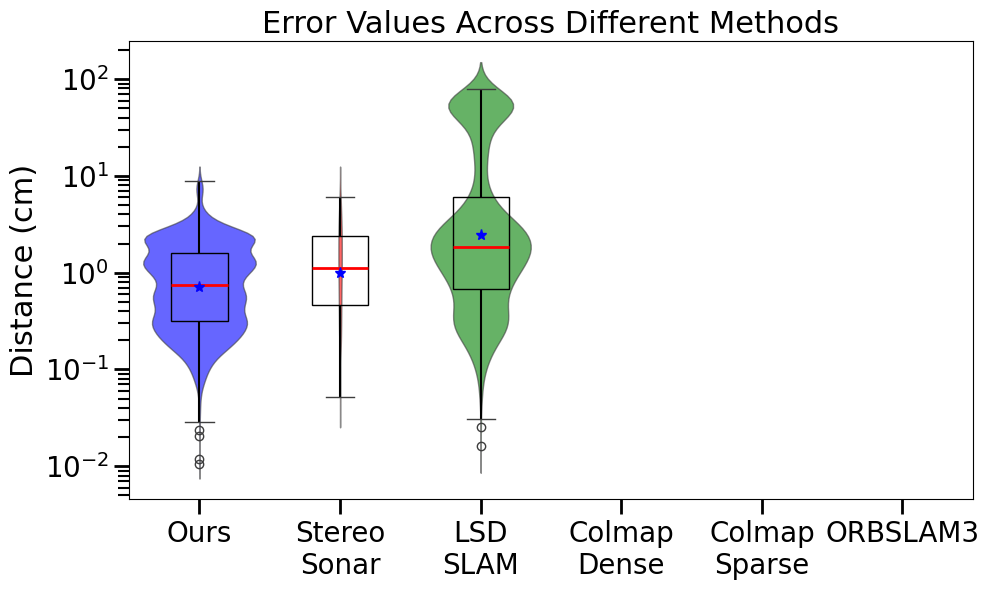} 
    \end{subfigure}
    \begin{subfigure}{0.24\linewidth}
        \centering
        \includegraphics[width=\linewidth]{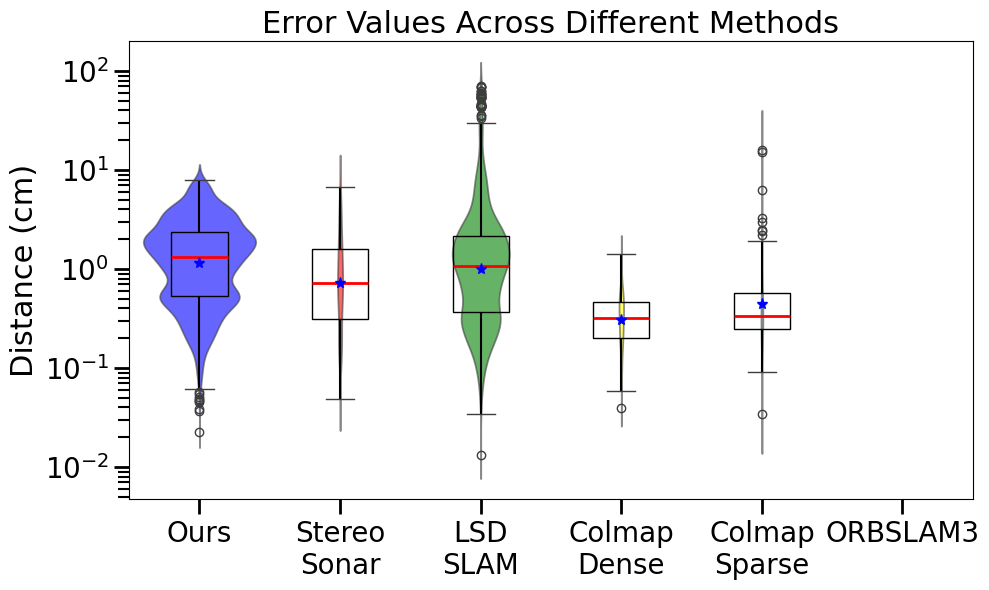} 
    \end{subfigure} \vspace{-1mm}
    \caption{\textbf{Tank pier pilings reconstruction results in different types of water.} Each column in the image shows results for a specific water type.
    The top row shows example images of original tank water (Type I) and other emulated water types (Type 5C, 7C and 9C).
    The middle row shows the ground-truth CAD model in gray, with different colored point clouds depicting each algorithm outputs. This row also shows the coverage of each algorithm expressed as voxel count. The bottom row shows absolute distance error values and error value distributions.}
    \label{fig:PierTankResults} \vspace{-6mm}
\end{figure*}

We assume that type-I water parameters are already present in the original underwater tank imagery. Thus, these attenuation coefficients were reduced from the type-5C, type-7C, and type-9C parameters used to simulate mild to highly turbid water. Examples of the original images (denoted as type-I) and the simulated turbidity results are shown in Figs. \ref{fig:PierTankResults} and \ref{fig:SeaWallTankResults}.
\vspace{-1mm}
\subsection{Indoor Tank Experiments}
Two different arrangements of structures were designed to resemble relevant real-world scenarios. The first setup is a mock-up of a pier composed of four cylindrical pilings. In the second setup, the pilings were placed against the tank wall in order to simulate a corrugated seawall. When collecting data, our ROV traverses the scene from starboard to port. The sonar was set to a 3m maximum range. 

The purpose of these experiments was to evaluate the performance of our approach under different turbidity conditions and compare against other state-of-the-art frameworks while having access to the ground truth geometry of the scene. Error metrics were computed by comparing the generated point clouds to a ground-truth CAD model of the environments. Absolute error is computed by finding the shortest distance between a given point in the final point cloud and the CAD model of the environment. The resulting point clouds first were manually aligned, then re-aligned using Iterative Closest Point (ICP), and in the cases of vision-based methods \cite{Campos2021, Engel2014, schoenberger2016}, manually scaled to compensate for scale ambiguity.

\subsubsection{\textbf{Tank Pier Results}}
A view of the CAD model and resulting point cloud reconstructions depending on water type can be observed in the second row of Fig. \ref{fig:PierTankResults}. This row also shows the coverage of each algorithm expressed as voxel count. The third row in the figure shows the reconstruction error distance and error distributions. Although COLMAP shows good performance in some water types such as type-I and type-5C, it performs poorly in other water types such as type-9C, and stops working completely in type-7C water. ORBSLAM3 shows overall poor performance in all trials. The Stereo Sonar approach has acceptable error performance in general; however, the coverage achieved by this method is low compared to others. LSD SLAM achieves good coverage; nevertheless, its error distributions skew higher than the error distributions obtained by our approach. Our approach shows stable performance in all water types, with its coverage increasing relative to other methods as turbidity increases. 

\begin{figure*}[tbh]
    \centering
    \begin{subfigure}{0.24\linewidth}
        \centering
        \includegraphics[width=\linewidth]{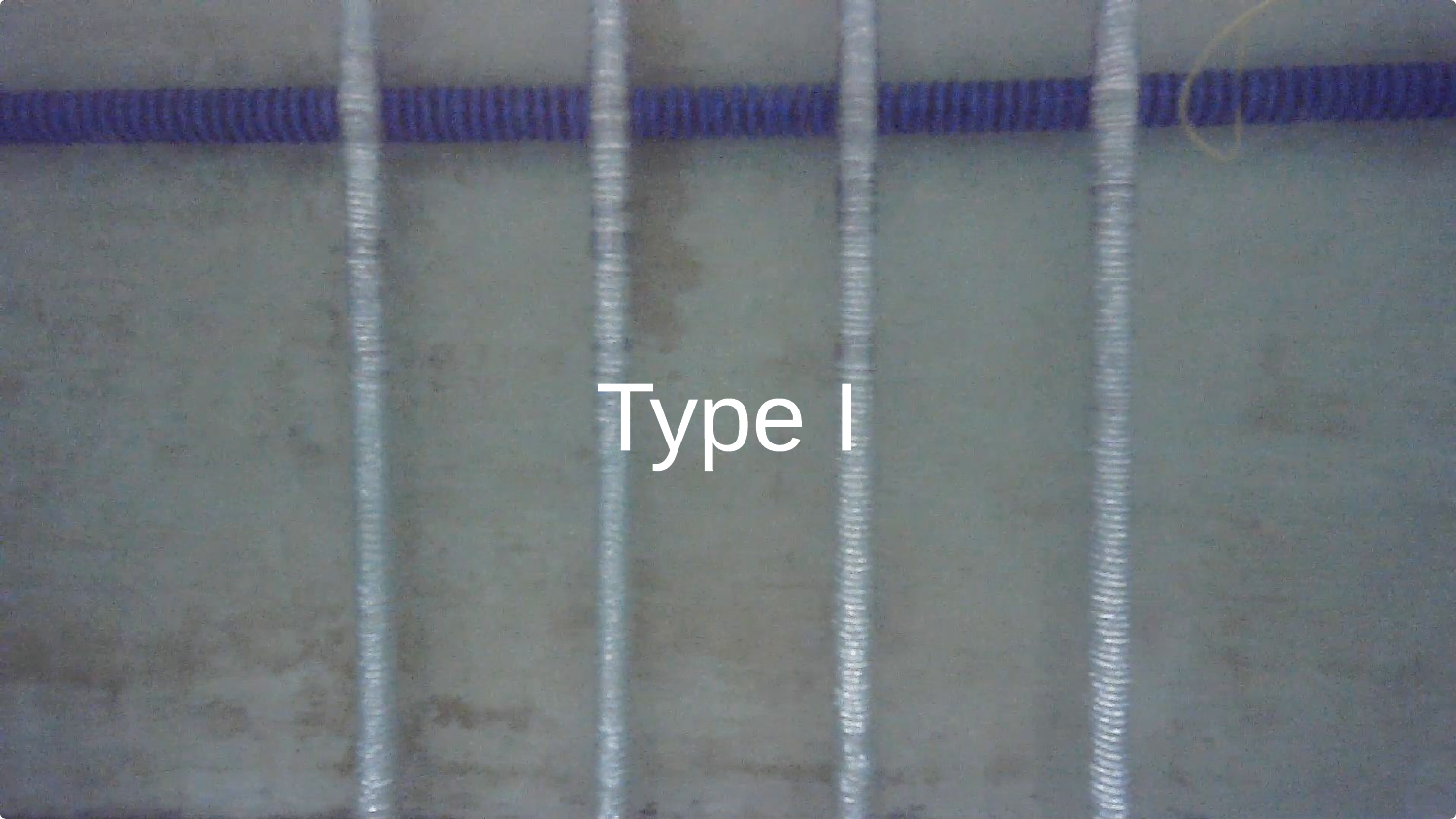} 
    \end{subfigure}
    \begin{subfigure}{0.24\linewidth}
        \centering
        \includegraphics[width=\linewidth]{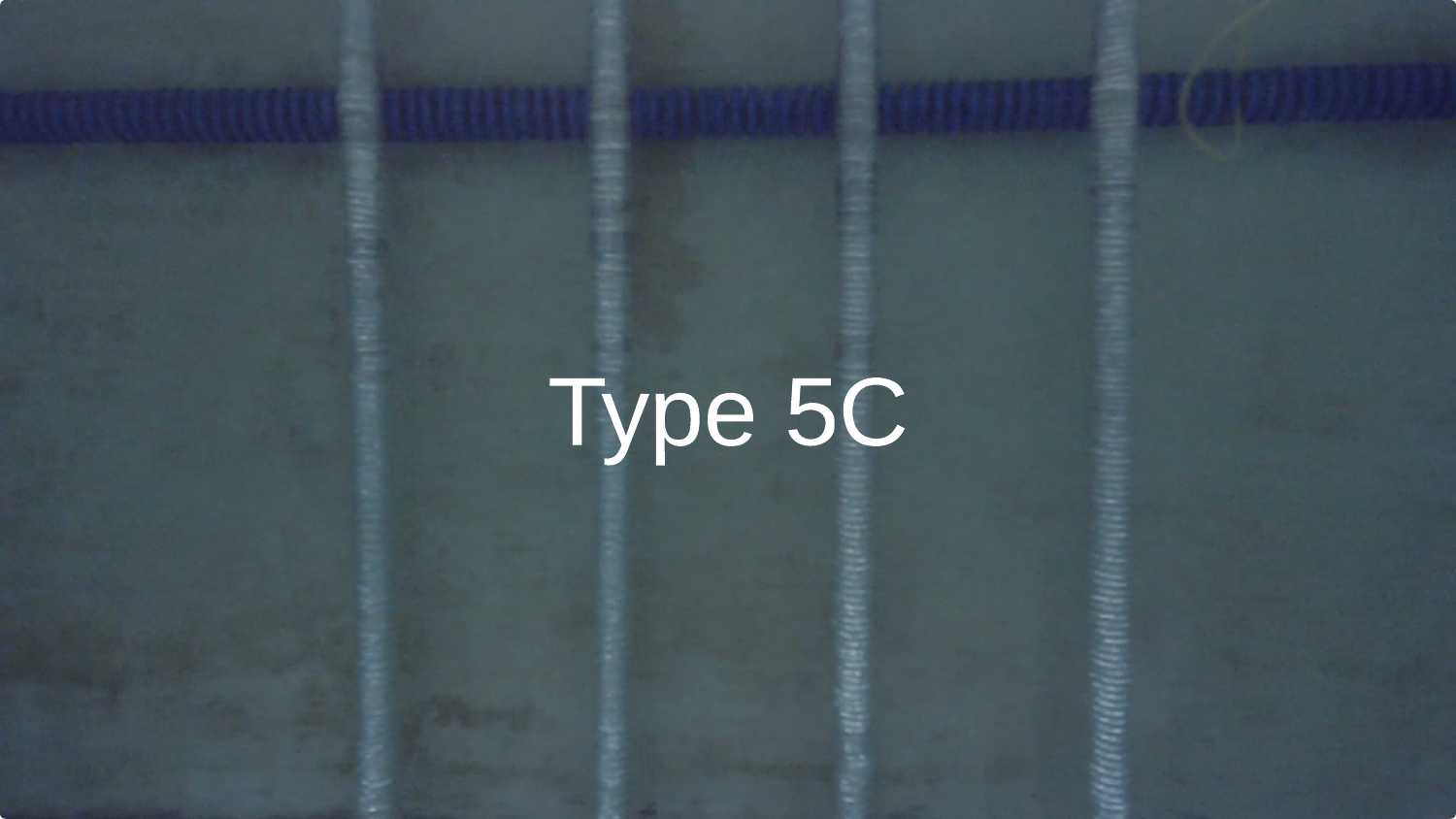} 
    \end{subfigure}
    \vspace{0.1cm}
    \begin{subfigure}{0.24\linewidth}
        \centering
        \includegraphics[width=\linewidth]{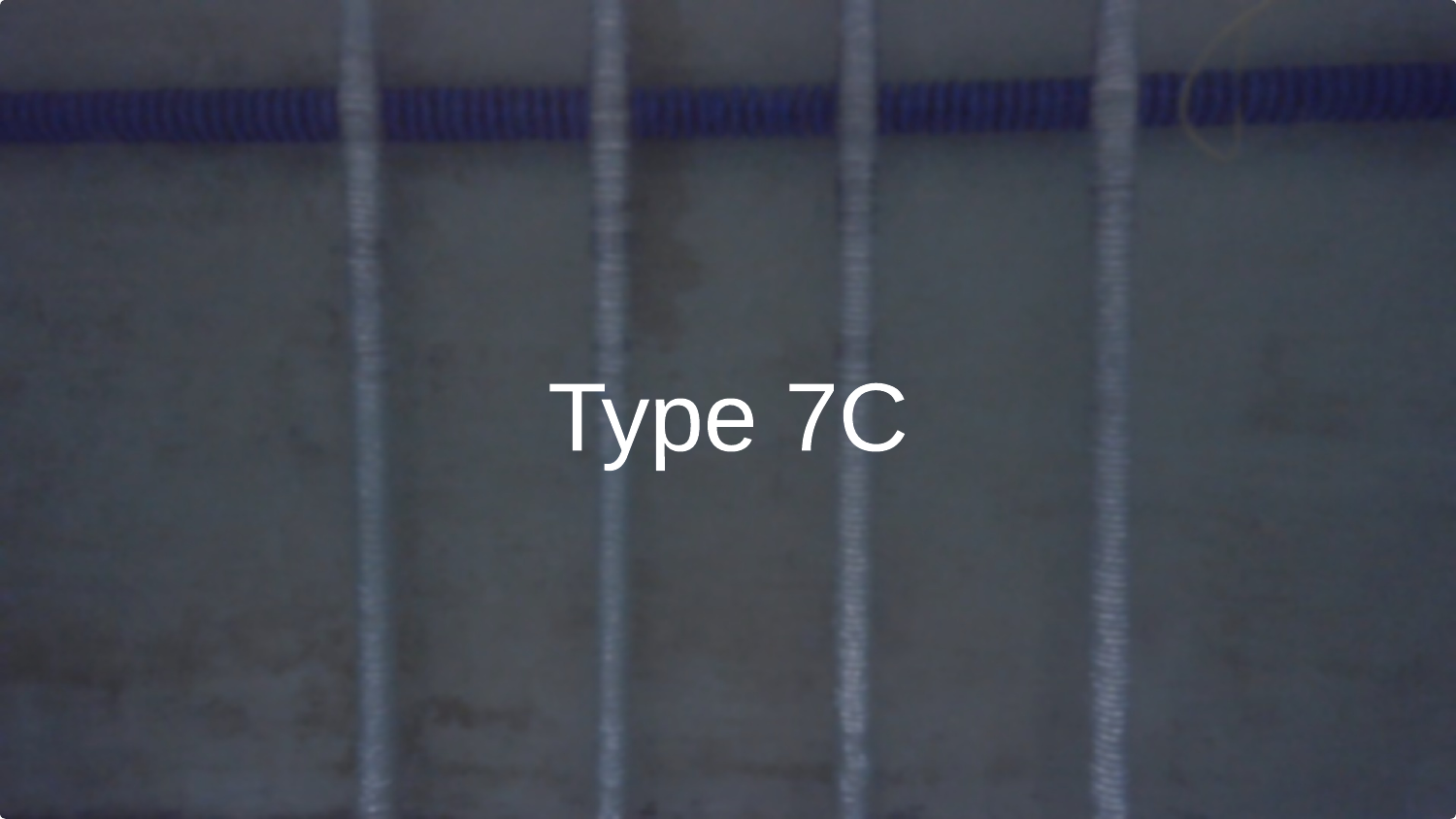} 
    \end{subfigure}
    \begin{subfigure}{0.24\linewidth}
        \centering
        \includegraphics[width=\linewidth]{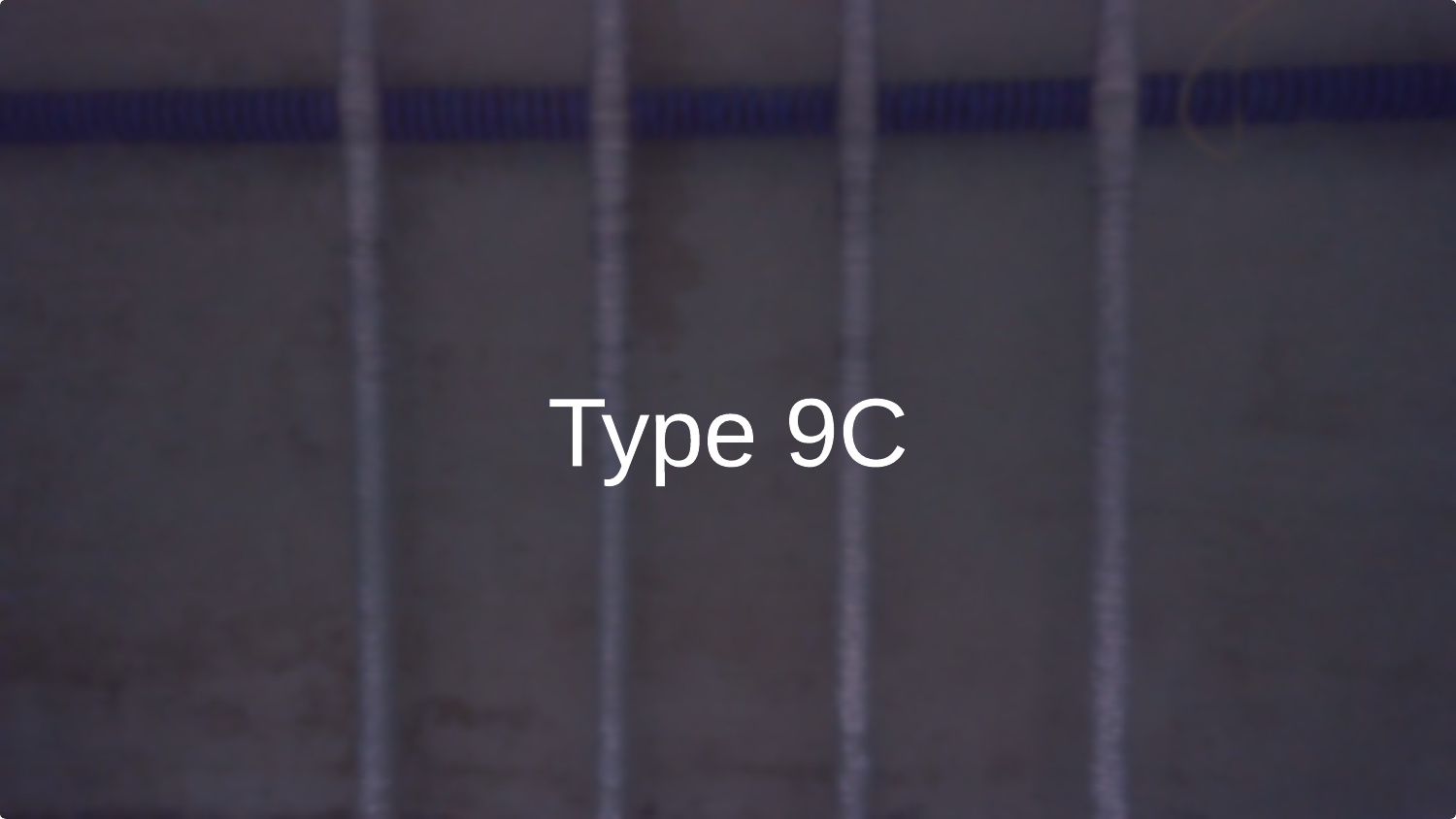} 
    \end{subfigure}
    \begin{subfigure}{0.24\linewidth}
        \includegraphics[width=\linewidth]{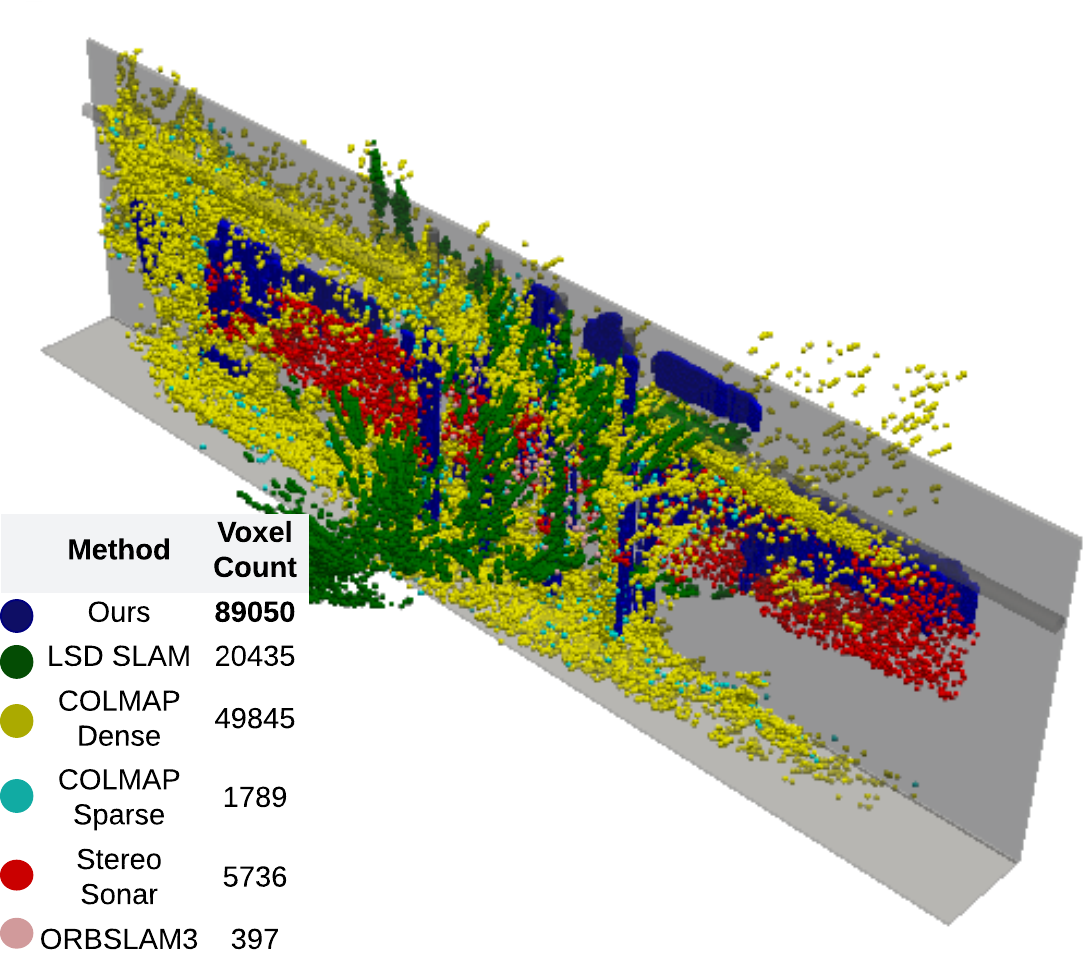} 
    \end{subfigure}
    \begin{subfigure}{0.24\linewidth}
        \centering
        \includegraphics[width=\linewidth]{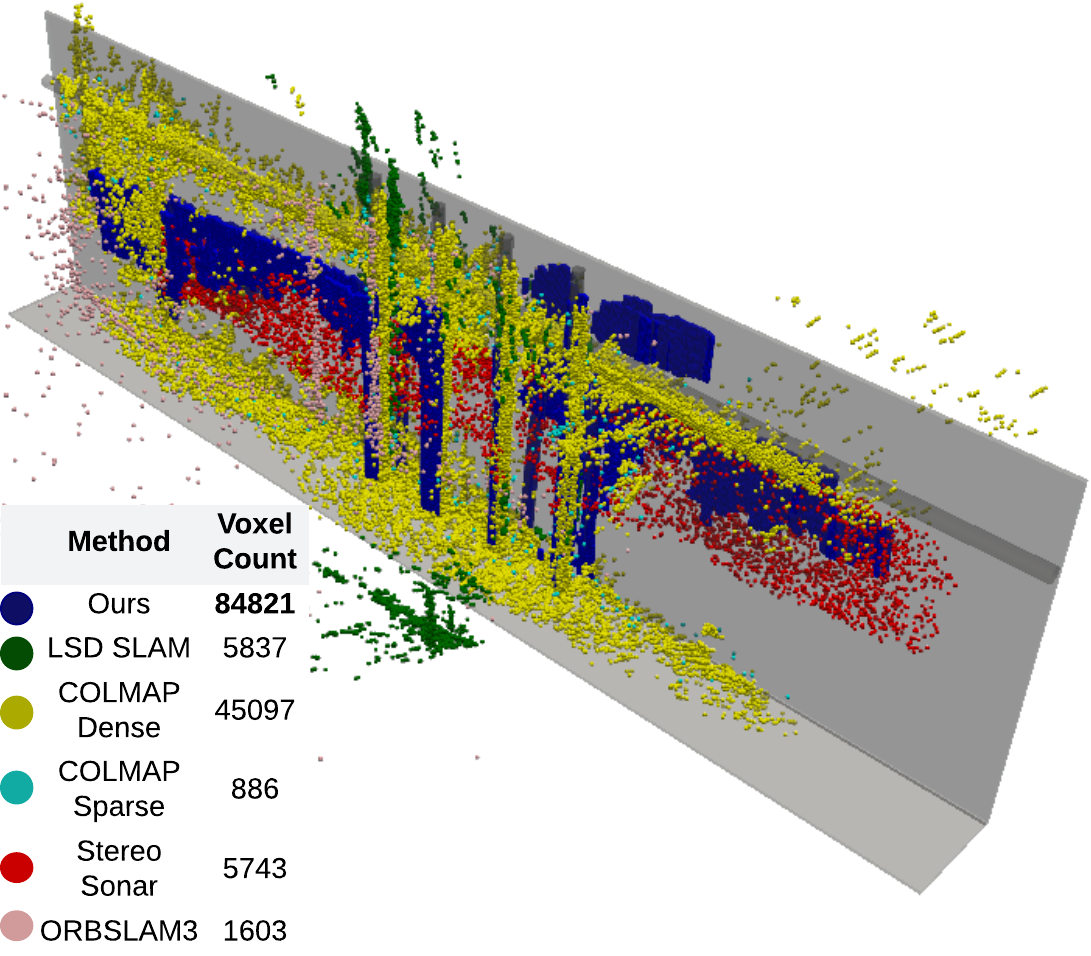} 
    \end{subfigure}
    \begin{subfigure}{0.24\linewidth}
        \centering
        \includegraphics[width=\linewidth]{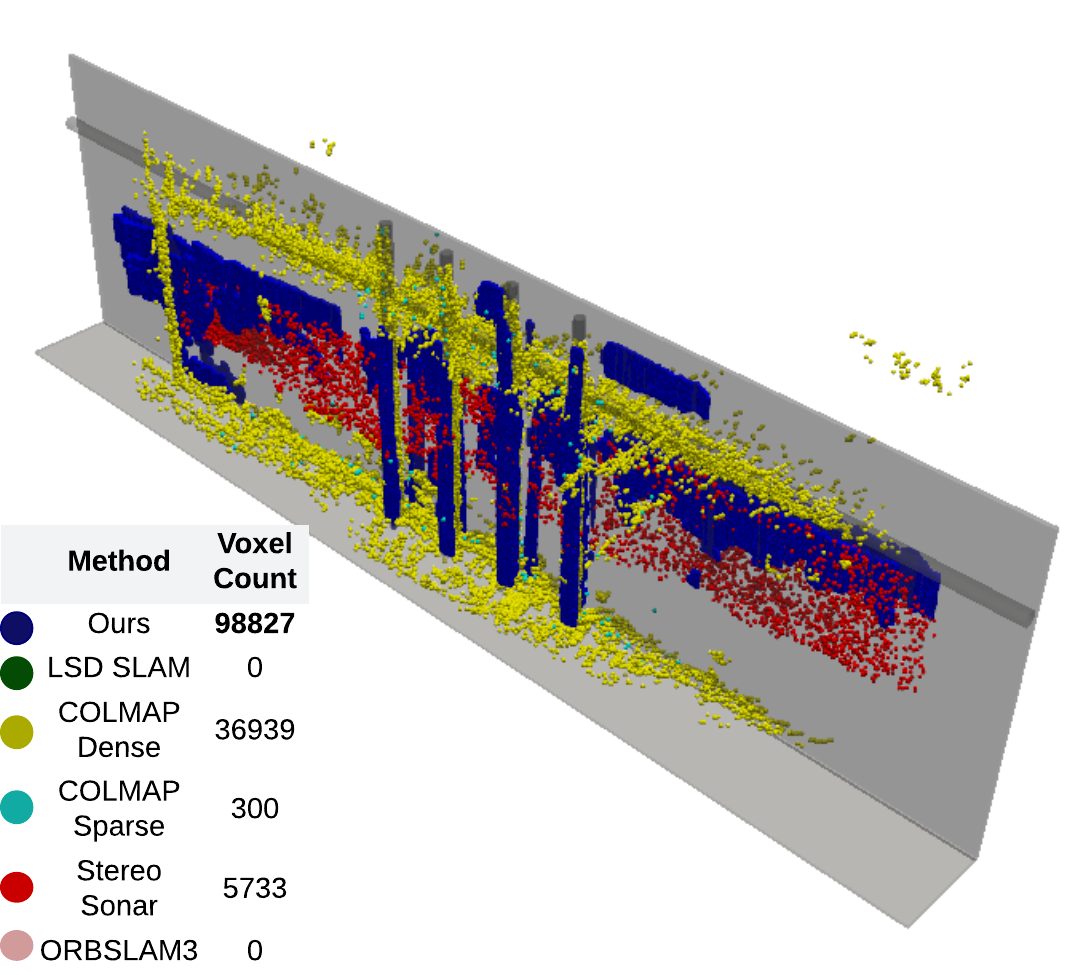} 
    \end{subfigure}
    \begin{subfigure}{0.24\linewidth}
        \centering
        \includegraphics[width=\linewidth]{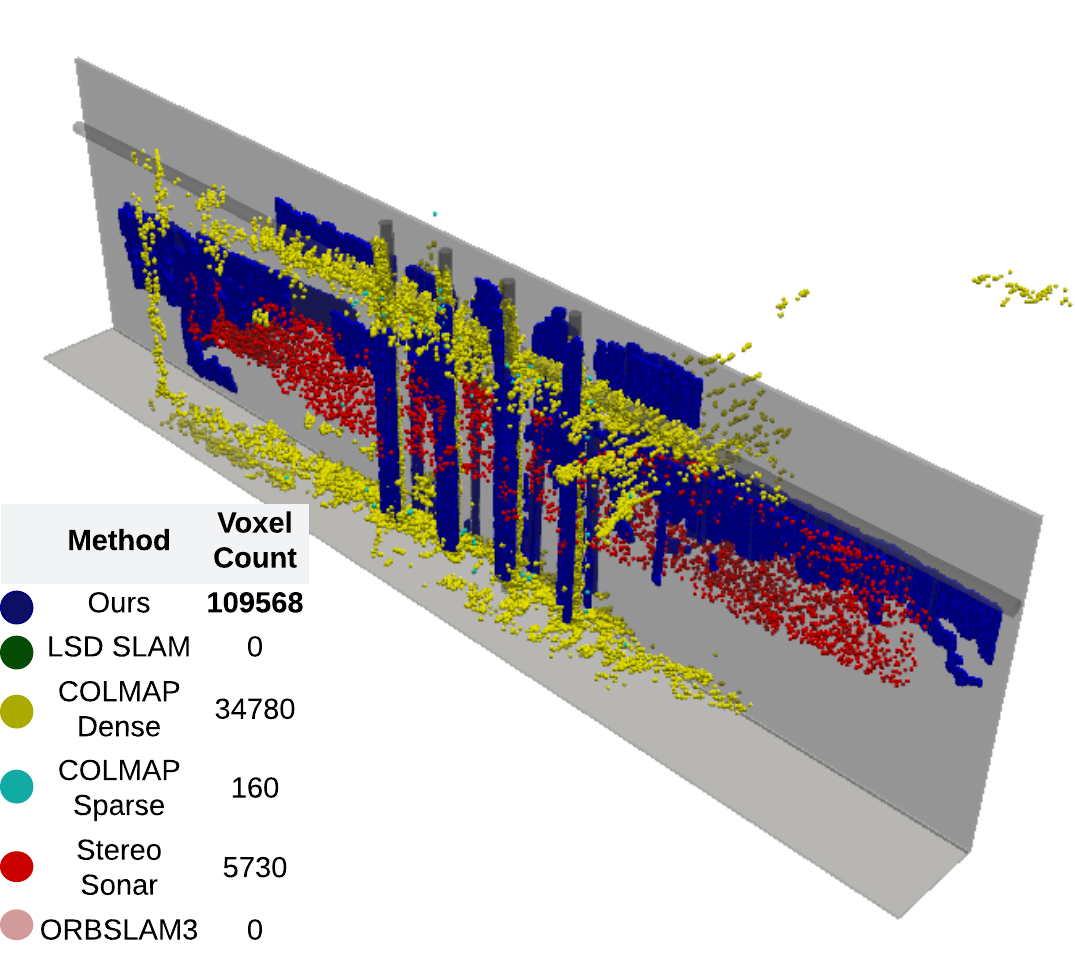} 
    \end{subfigure}
    \begin{subfigure}{0.24\linewidth}
        \centering
        \includegraphics[width=\linewidth]{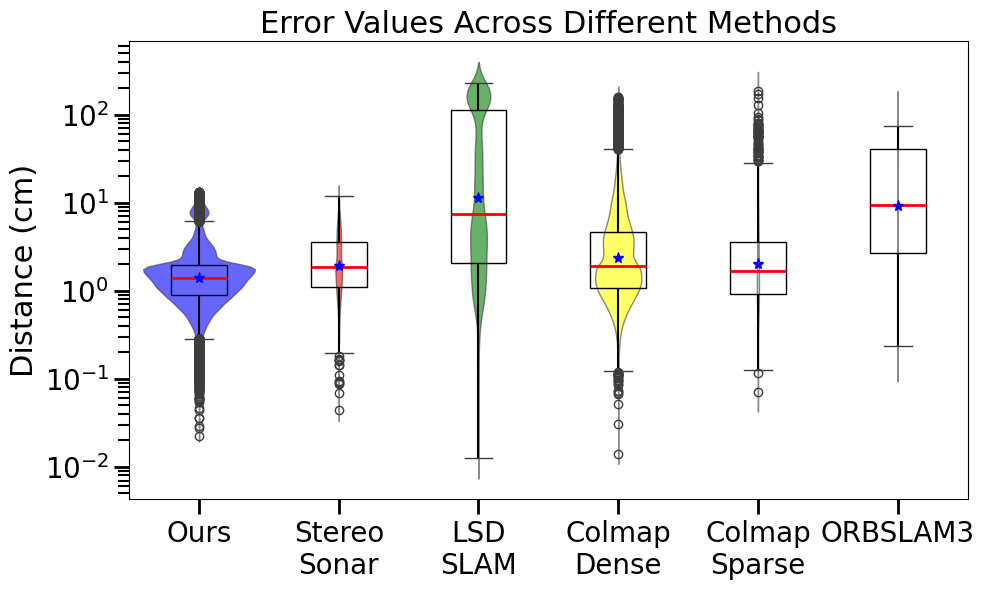} 
    \end{subfigure}
    \begin{subfigure}{0.24\linewidth}
        \centering
        \includegraphics[width=\linewidth]{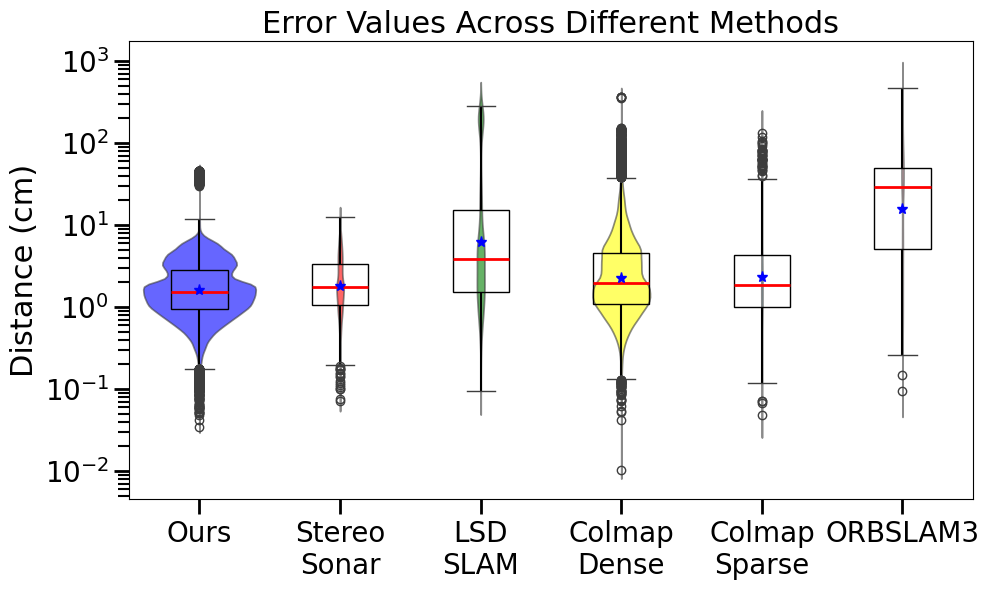} 
    \end{subfigure}
    \begin{subfigure}{0.24\linewidth}
        \centering
        \includegraphics[width=\linewidth]{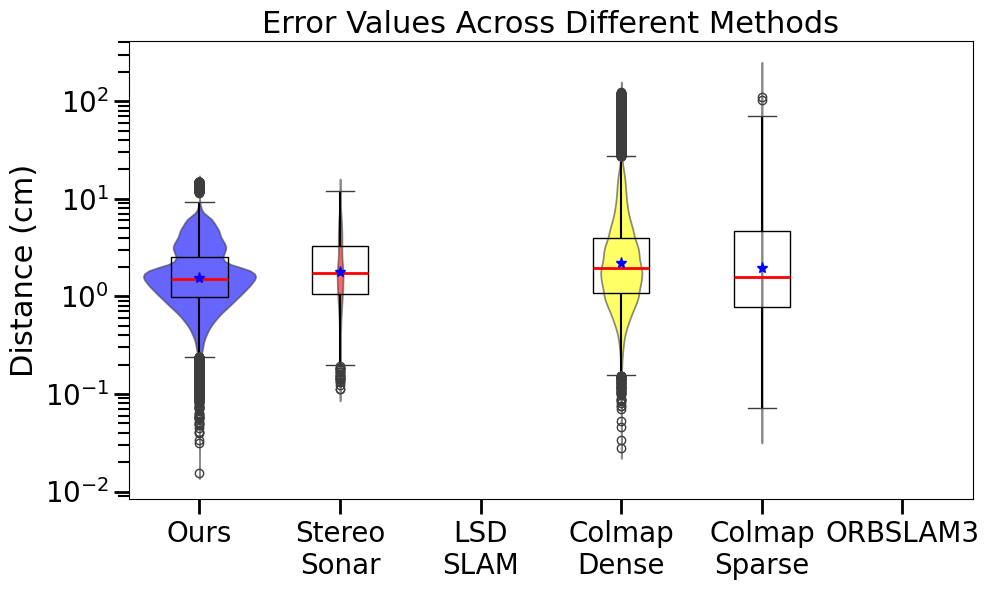} 
    \end{subfigure}
    \begin{subfigure}{0.24\linewidth}
        \centering
        \includegraphics[width=\linewidth]{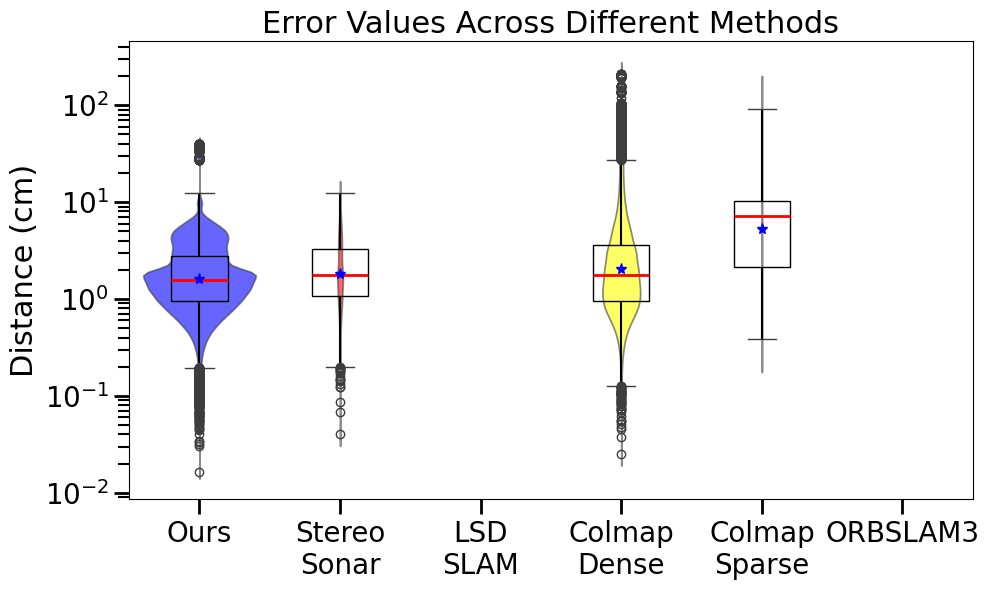} 
    \end{subfigure}
    \vspace{-1mm}
    \caption{\textbf{Tank sea wall reconstruction results in different types of water.} Each column in the image shows results for a specific water type.
    The top row shows example images of original tank water (Type I) and other emulated water types (Type 5C, 7C and 9C).
    The middle row shows the ground-truth CAD model in gray, with different colored point clouds depicting each algorithm outputs. This row also shows the coverage of each algorithm expressed as voxel count. The bottom row shows absolute distance error values and error value distributions.}
    \label{fig:SeaWallTankResults} \vspace{-6mm}
\end{figure*}
\subsubsection{\textbf{Tank Sea Wall Results}}
Fig. \ref{fig:SeaWallTankResults} shows the resulting point cloud reconstructions along with coverage, the error distance values, and distributions per method and water type. Again, the Stereo Sonar method shows stable performance, but its coverage is low compared to visual methods. COLMAP achieves acceptable error, however the point cloud reconstructions in the top row of Fig.\ref{fig:SeaWallTankResults} show failure to map the smooth wall surface, primarily capturing the pilings, lane markers, and tank floor, which provide better visual features. ORBSLAM3 and LSD SLAM completely stop working in type-7C and type-9C water types. It is evident that our method consistently outperforms others, demonstrating lower-than-average error across various water types. Furthermore, our method provides superior coverage results compared to alternative approaches. These findings highlight that our method is more effective at handling feature-sparse visual data, particularly when compared to other optical methods that rely on point features.

\begin{figure}[tbh] 
    \centering
    \begin{subfigure}{0.31 \columnwidth}
        \centering
    \includegraphics[width=\linewidth]{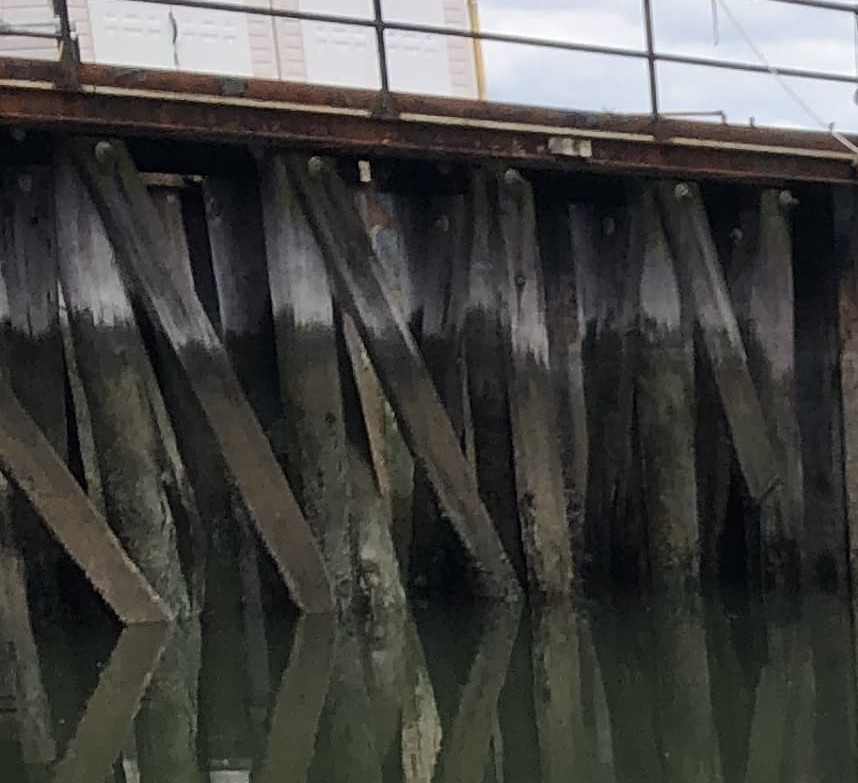} 
        \label{fig:PierField}
    \end{subfigure}
    \begin{subfigure}{0.3\columnwidth}
        \centering
        \includegraphics[width=\linewidth]{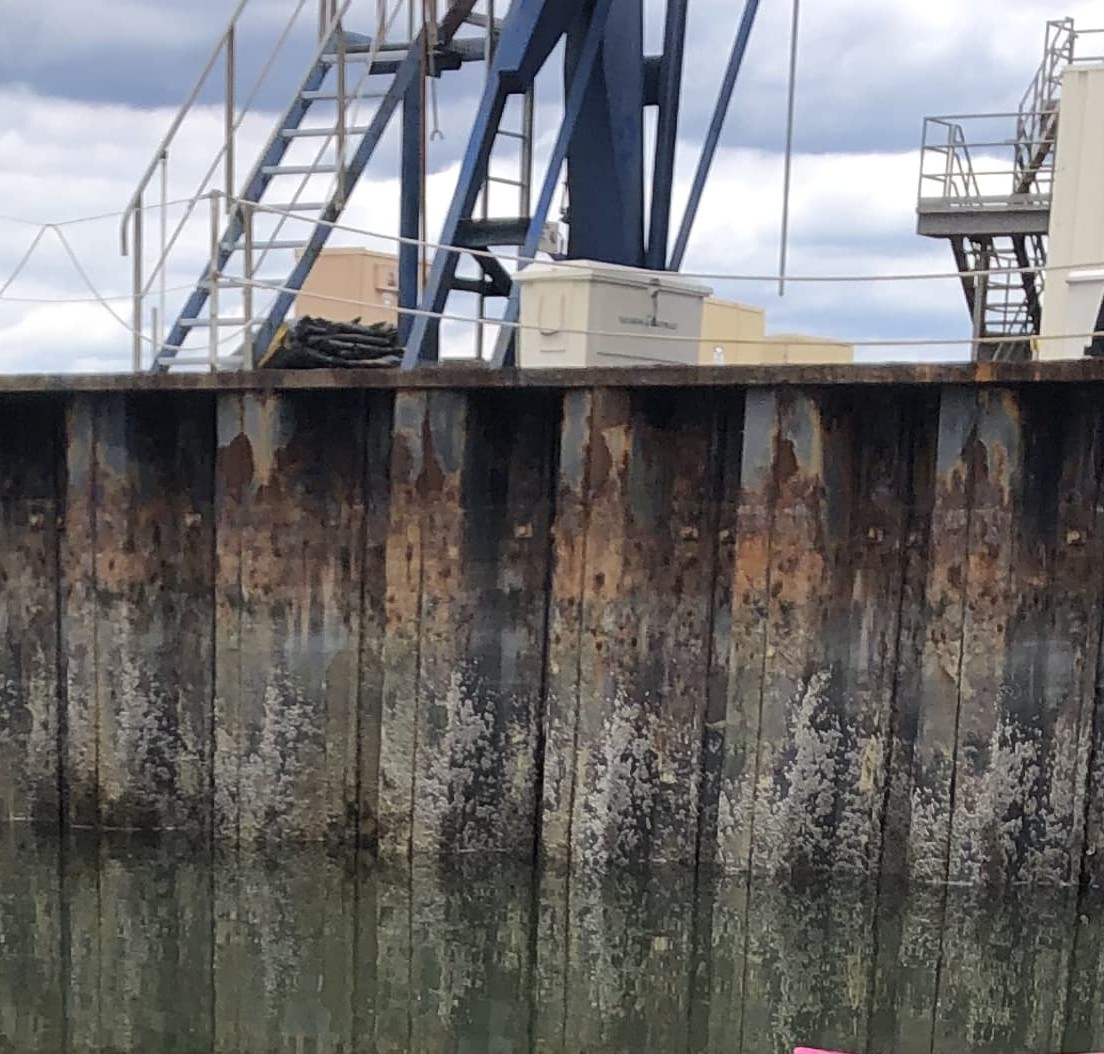} 
        \label{fig:SeaWallField}
    \end{subfigure} \vspace{-4mm}
    \caption{\textbf{Field environments.} Pier (left) and corrugated seawall (right).} \vspace{-5mm}
    \label{fig:FieldStructures}
\end{figure} 
\vspace{-1mm}
\subsection{Outdoor Field Experiments}
In order to demonstrate the performance of our proposed approach in the field, we deployed our ROV at the United States Merchant Marine Academy (USMMA) marina, located in King's Point, NY. This environment exhibits a high level of turbidity, as can be seen in the example image in Fig. \ref{fig:method}. The vision-based approaches including LSD SLAM \cite{Engel2014}, ORBSLAM3 \cite{Campos2021}, and COLMAP \cite{schoenberger2016}, used during the tank experiment were not capable of finding sufficient features in the optical imagery to provide a resulting reconstruction. Thus, our approach is only compared against the Stereo Sonar \cite{McConnell2020} reconstruction method. 
Two separate structures were observed (shown in Fig. \ref{fig:FieldStructures}), a pier with wooden pilings, and a concrete corrugated sea wall. Because of the lack of ground truth data of these structures, only the coverage of each algorithm can be numerically compared; the results are shown in Table \ref{tab:FieldResultsCoverage}. 
\begin{table}[ht] \vspace{-2mm}
\centering
\begin{tabular}{|l|c|c|}
\hline
\multicolumn{3}{|c|}{ \textbf{Voxel Count}}\\
 \hline
 \textbf{Structure} & \textbf{Ours} & \textbf{Stereo Sonar} \\
\hline
Pier & \textbf{24956} & 3867 \\
\hline
Sea Wall & \textbf{7412} & 851 \\
\hline
\end{tabular}
\caption{\textbf{Field coverage results}, where point clouds are voxelized using a 1cm grid cell resolution. Corresponding structures are shown in Fig. \ref{fig:FieldStructures}.} \vspace{-2mm}
\label{tab:FieldResultsCoverage}
\end{table}
\subsubsection{\textbf{Pier Results}}
\begin{figure}[tb]
\centerline{\includegraphics[width=0.5\textwidth]{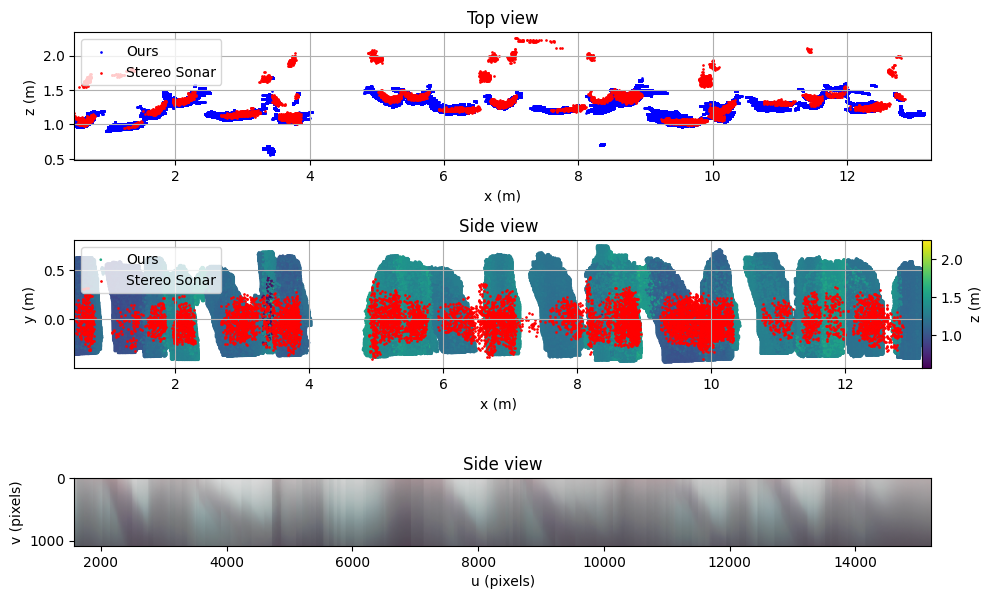}} \vspace{-2mm}
\caption{\textbf{Pier field experiment results.} Reconstruction results are shown in the top two rows. The bottom row shows unscaled optical data.}
\label{fig:PierFieldViews} \vspace{-6mm}
\end{figure}

In this experiment, the vehicle traverses from port to starboard and the sonar was operated at a 3m maximum range. 
The top and side views of the resulting pier point cloud reconstructions can be seen in Fig. \ref{fig:PierFieldViews}. The top row shows points from the Stereo Sonar method in red and points from our proposed method in blue. In the second row, the Stereo Sonar result is again shown in red, however results from our method are shown using a colored distance scale. The bottom row of this figure shows a visual representation of the optical data; this mosaic representation was manually constructed and is not scaled to fit the ground truth metric data, it is only meant to give the reader an example of the visual field data. 

It can be observed from Table \ref{tab:FieldResultsCoverage} that the coverage of our proposed method is superior to the coverage achieved by the Stereo Sonar approach. It is evident from the top view in Fig. \ref{fig:PierFieldViews} that our approach was capable of reconstructing two small pipe structures that the Stereo Sonar approach completely failed to map, likely due to their small size and close proximity to the robot. Lastly, the pier structure contained several diagonal beams; it is impossible to discern the diagonal shape in the Stereo Sonar reconstruction, and in contrast, our method is capable of capturing this changing cross-section. 

\subsubsection{\textbf{Corrugated Seawall Results}}
\begin{figure}[tb] \vspace{-2mm}
\centerline{\includegraphics[width=0.5\textwidth]{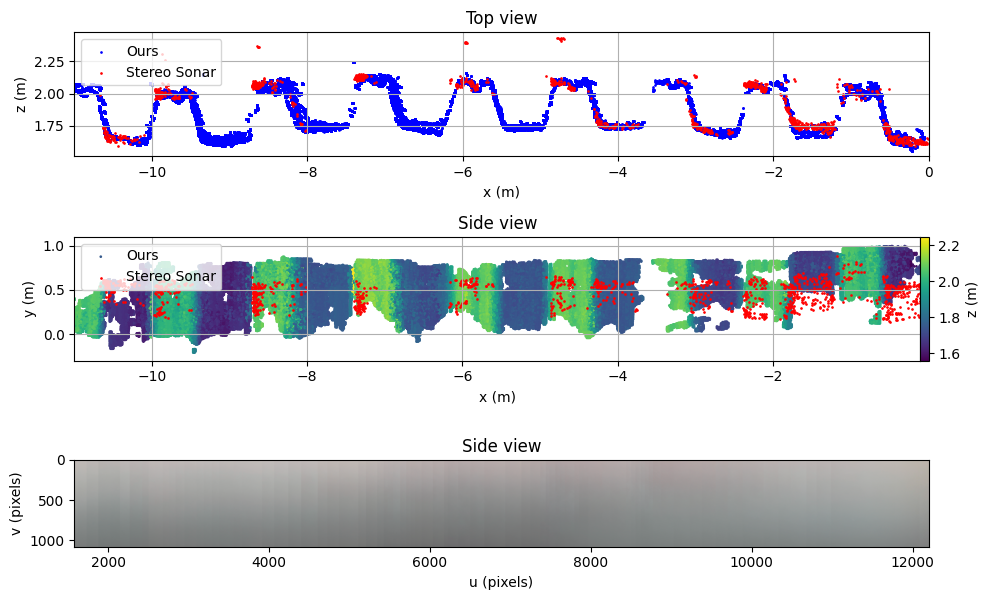}} \vspace{-3mm}
\caption{\textbf{Seawall field experiment results.} Reconstruction results are shown in the top two rows. The bottom row shows unscaled optical data.}
\label{fig:SeaWallFieldViews} \vspace{-6mm}
\end{figure}
In this experiment, the vehicle traversed the scene from starboard to port, and the sonar was operated at a 5m maximum range. The top and side views of the resulting seawall reconstruction can be seen in Fig. \ref{fig:SeaWallFieldViews}. It is evident from the coverage results in Table \ref{tab:FieldResultsCoverage} that our proposed approach achieves superior coverage compared to the Stereo Sonar approach. It is also clear from Fig. \ref{fig:SeaWallFieldViews} that our method is capable of recreating the corrugated wall geometry, while expanding the vertical FOV. 

\vspace{-1mm}
\section{Conclusion}
\vspace{-1mm}
This paper introduces an opti-acoustic scene reconstruction method designed for highly turbid underwater environments. The approach begins with image object segmentation, followed by the projection of sonar beams onto a 2D image. Next, sonar-to-image matching is performed, and sonar-derived distances are fused with image elevation data to reconstruct the scene. Unlike traditional methods, our approach avoids point-based feature detection in visual images, which is unreliable in turbid conditions. Our method operates in real-time and can perform reconstructions from a single pose, eliminating the need for multi-view data. 
We evaluate its performance against state-of-the-art techniques across various turbidity levels, demonstrating comparable accuracy while maintaining robust coverage of reconstructed environments. In addition, field tests highlight our method’s capability to recreate human-made structures in highly turbid settings. Our method's simplified approach does not require training and contains few tuned parameters, while being applicable in diverse settings, including shallow water settings where sunlight effects can vary substantially. This approach is meant to aid navigation towards and perception of close-range objects needed for manipulation tasks in turbid water environments. \color{black}This method performs well in man-made environments with varying object cross-sections, but its assumption of constant distance 
above and below each sonar range return limits effectiveness in scenes with complex geometries. 
\color{black}

\vspace{-2mm}
\bibliographystyle{IEEEtran}
\vspace{-4mm}
\bibliography{bib.bib}

\end{document}